\documentclass{tlp}
\usepackage{aopmath}

\usepackage{graphicx}

\usepackage{graphicx}
\usepackage{algorithmic}
\usepackage{url}
\usepackage{natbib}

\usepackage{subfigure}

\newcommand{\comment}[1]{}

\newtheorem{definition}{Definition} 

\begin{document}

\long\def\comment#1{}

\submitted{30 December 2014}
\revised{13 July 2015}
\accepted{14 December 2015}

\title[An Event Calculus Production Rule System]{An Event Calculus Production Rule System for Reasoning in Dynamic and Uncertain Domains}

\author[T. Patkos et al.]
{Theodore Patkos, Dimitris Plexousakis\\
Institute of Computer Science, FO.R.T.H.\\
E-mail: \{patkos, dp\}@ics.forth.gr
\and
Abdelghani Chibani, Yacine Amirat \\
Lissi Laboratory, Paris-Est Cr\'eteil\\
Val-de-Marne University (UPEC)\\
E-mail: \{chibani,amirat\}@u-pec.fr
}

\pagerange{\pageref{firstpage}--\pageref{lastpage}}
\volume{\textbf{10} (3):}
\jdate{December 2014}
\setcounter{page}{1}
\pubyear{2016}

\maketitle

\label{firstpage}

\begin{abstract}
Action languages have emerged as an important field of Knowledge Representation
for reasoning about change and causality in dynamic domains. This article presents Cerbere, a production system designed to perform online causal, temporal and epistemic reasoning based on the Event Calculus. The framework implements the declarative semantics of the underlying logic theories in a forward-chaining rule-based reasoning
system, coupling the high expressiveness of its formalisms with
the efficiency of rule-based  systems. To illustrate its applicability, we present both the modeling of benchmark problems in the field, as well as its utilization in the challenging domain of smart spaces. A hybrid framework that combines logic-based with probabilistic reasoning has been developed, that aims to accommodate activity recognition and monitoring tasks in smart spaces.\\
\textit{Paper under consideration for publication in the Theory and Practice of Logic Programming.}
\end{abstract}
\begin{keywords}
Event Calculus, Rule-based Reasoning, Smart Environments
\end{keywords}

\section{Introduction}
\label{intro}

Reasoning about actions, change and causality has been an important challenge from the early days of Artificial Intelligence (AI). The Event Calculus \citep{KowalskiSergot86,Miller02}, a well-established technique for reasoning about causal and narrative information in dynamic environments, has been applied in domains as diverse as high-level robot cognition, argumentation,
 service composition,
 complex event detection, 
and others. 

Satisfiability- and logic programming-based implementations of Event Calculus dialects have been proposed over the years. Recently, progress in generalizing the definition of stable model semantics used in Answer Set Programming (ASP) \citep{FerrarisSMC11} has opened the way for the reformulation of Event Calculus axiomatizations into logic programs that can be executed with ASP solvers \citep{LeeRSC12}. Moreover, powerful extensions of the main formalism have been developed to accommodate, for instance, probabilistic uncertainty \citep{skarlTPLP15} or knowledge derivations with non-binary-valued fluents \citep{MillerMP13}. Due to its elegance in carrying out reasoning tasks, such as deductive narrative verification and abductive planning, most popular implementations to date rely on a backward-chaining style of computation, which is mainly goal-driven. 

Recently, significant emphasis has been given towards real-time computing both in research and in industry, which calls for efficient, reactive systems. Runtime reasoning tasks, such as monitoring, can greatly benefit from the Event Calculus style of domain representation, yet logic-based systems are in general not optimized for run-time event processing \citep{AnicicFodor10}. Forward-chaining systems are particularly efficient in implementing such a formalism, in order to dynamically react to streams of occurring events \citep{Bragaglia12}. Yet, embedding declarative reasoning rules in a production system is not a trivial task, especially when complex commonsense features are to be supported. Most Event Calculus-like reactive implementations often dismiss their declarative nature, which can prove problematic in terms of preserving the formal properties of the underlying theories, as argued by \cite{ChesaniMelloMontali10}.  

In this paper, we present the design and implementation of a production system for causal, temporal and epistemic reasoning that aims to contribute towards filling this gap.\footnote{This article is partially based and extends the research presented in \citep{PatkosRuleML12}. Executables and sample axiomatizations mentioned in the text can be found in the Appendix accessible online at \url{http://www.csd.uoc.gr/~patkos/tplpAppendix/TPLP16Appendix.pdf}.} In particular:
\begin{itemize}
\item we describe Cerbere (Causal and Epistemic Rule-Based Event calculus REasoner), a system that translates an expressive Event Calculus variant into a rule-based program, capable of reacting to occurring events using the production rules paradigm. Although online reasoners for the Event Calculus have emerged recently, they usually implement expressively restricted variants of the Event Calculus, and to our knowledge no existing tool can accommodate triggered events, non-determinism, state constraints and reasoning about knowledge in a production framework.

\item our system aims to transfer the benefits of the underlying formalisms, such as the solution to the frame problem for expressive classes of problems, into an efficient forward-chaining system that goes beyond ordinary rule-based systems deployed in dynamic domains, where the actions that lead to the assertion and retraction of facts have no real semantics and high-level structures. Instead, it uses the structures of the formal theories to define the causal properties of actions or to manipulate ordinary and epistemic context-dependent facts. A repertoire of technical solutions has been developed, ranging from mechanisms for managing the size of the Knowledge Base (KB) to intuitive graphical user interfaces, in order to enhance its performance.

\item the applicability of the proposed system is illustrated in the challenging field of smart spaces. Specifically, we present how the reasoner can be seamlessly coupled with other components in a framework that can perform activity recognition, monitoring and action execution tasks. An extension of a commonly used methodology for probabilistic reasoning, based on Bayesian Networks (BNs), is integrated with our logic-based reasoner, in order to present the benefits that can be achieved, both with respect to the breadth of phenomena that can be modeled and the efficiency of reasoning. The evaluation analysis performed verifies its applicability to such domains.
\end{itemize}

Of course, Cerbere is not aimed as a holistic solution for the inference problems encountered in smart spaces, or dynamic systems in general. While production systems, such as ours, are well-suited for implementing run-time reactive behavior, decision making and planning are better supported by systems relying on logic programming. Towards this end, for instance, \cite{KowalskiSadri15,KowalskiSadri10,KowalskiSadri96} propose a language that aims to reconcile backward with forward reasoning inside an intelligent agent architecture. Still, the plurality of features supported by Cerbere significantly extend the domains that can by accommodated by other similar systems.

The paper proceeds with 6 main sections. Section \ref{background} introduces the Event Calculus, and Section \ref{sec3} describes our implementation of the formalism using a rule-based approach for online execution. Section 4 presents a hybrid framework for reasoning in smart spaces, describing how it couples logic-based reasoning with a probabilistic component. Section \ref{sec:experiments} evaluates the performance of the reasoner and Section \ref{discussion} discusses related platforms. The paper concludes in Section 7 with general remarks and future work.

\section{The Theoretical Foundations}
\label{background}


\label{DEC}

The Event Calculus is a narrative-based
many-sorted first-order language for reasoning about action and
change, where the sort $\cal{E}$ of \emph{events} indicates changes in the environment,
the sort $\cal{F}$ of \emph{fluents} denotes time-varying properties and the
sort $\cal{T}$ of \emph{timepoints} is used to implement a linear time structure. The calculus applies the
\emph{principle of inertia} for fluents, in order to solve the frame
problem, which captures the property that things
tend to persist over time unless affected by some event. It also relies on the technique of \emph{circumscription} \citep{Lifschitz1994c}  to support default reasoning. A set of predicates is
defined to express which fluents hold when ($HoldsAt \subseteq \cal{F} \times \cal{T}$), which events
happen ($Happens \subseteq \cal{E} \times \cal{T}$), which their effects are ($Initiates$,
$Terminates$, $Releases \subseteq \cal{E} \times \cal{F} \times \cal{T}$) and whether a fluent is subject to the law
of inertia or released from it ($ReleasedAt \subseteq \cal{F} \times \cal{T}$).\footnote{In the
sequel, variables, starting with a lower-case letter, are implicitly
universally quantified, unless otherwise stated. Predicates and constants start with an upper-case letter. Variables of the sort $\cal{E}$ are represented by $e$, fluent
variables by $f$ and variables of the sort $\cal{T}$ by $t$, with
subscripts where necessary.}

Our account of action and knowledge in this paper is formulated within the
circumscriptive linear Discrete time Event Calculus, extensively described
in \citep{Mueller06}. The commonsense notions of persistence and
causality are captured in a set of \emph{domain independent} axioms, referred to as $\mathcal{DEC}$, that express the influence of events
on fluents and the enforcement of inertia for the $HoldsAt$ and
$ReleasedAt$ predicates. In brief, $\mathcal{DEC}$ states that a fluent that is not released from inertia has a particular truth value at a particular time if at the previous timepoint either it was given a cause to take that value or it already had that value. For example, $Initiates(E,F,T)$ means that if action $E$ happens at timepoint $T$ it gives cause for fluent $F$ to be true at timepoint $T+1$. A fluent released from the law of inertia at a particular time may have a fluctuating truth value; this technique is used to introduce non-determinism to a domain axiomatization. 



\begin{table}[!t]
\renewcommand{\arraystretch}{1.3}
\caption{Event Calculus Types of Formulae}
\label{Domain axioms table}
\centering
\begin{tabular}{l|c}
\hline\hline
\bfseries Axiom Name & \bfseries Axiom\\
\hline
- $\bigwedge ^{f_i \in C_f} [HoldsAt(f_i,t)] \Rightarrow Initiates(E,F,t)$ & Positive Effect Axioms ($\Sigma$)\\
- $\bigwedge ^{f_i \in C_f} [HoldsAt(f_i,t)] \Rightarrow Terminates(E,F,t)$ & Negative Effect Axioms ($\Sigma$)\\
- $\bigwedge ^{f_i \in C_f} [HoldsAt(f_i,t)] \Rightarrow Releases(E,F,t)$ & Release Axioms ($\Sigma$)\\
- $\bigwedge ^{f_i \in C_f} [HoldsAt(f_i,t)] \Rightarrow HoldsAt(F,t)$ & State Constraints ($\Psi$)\\
- $\bigwedge ^{f_i \in C_f} [HoldsAt(f_i,t)] \wedge \bigwedge ^{e_i \in C_e}
[Happens(e_i,t)] \Rightarrow$ & Trigger Axioms ($\Delta_2$)\\
 \hspace*{0.6em} $Happens(E,t)$& \\
- $HoldsAt(F,T)$ or $ReleasedAt(F,T)$ & Observations ($\Gamma$)\\
- $Happens(E,T)$ & Event Occurrences ($\Delta_1$)\\
\hline\hline
\end{tabular}
\end{table}

In addition to domain independent axioms, a particular domain axiomatization requires also axioms that describe the
commonsense domain of interest, observations of world properties at
various times and a narrative of known world events. Table \ref{Domain axioms table} summarizes the main types of axioms
that can be used to describe a domain, where sets $C_f$ and $C_e$ denote the context of an axiom (precondition fluents and events, respectively), i.e. $C_f=\{F_1,...,F_n\}$, $C_e=\{E_1,...,E_m\}$, with $n,m\geq 0$. Formally, a domain description is defined as follows:

\begin{definition}[Event Calculus Domain Description]
An Event Calculus domain description $\mathcal{D}=
\langle\Sigma, \Delta_2, \Psi, \Gamma, \Delta_1, \Omega\rangle$
consists of:\\
- a set $\Sigma$ of \emph{positive effect}, \emph{negative effect} and
\emph{release axioms} that describe conditional effects of actions,\\
- a set $\Psi$ of \emph{state constraints},\\
- a set $\Delta _2$ of \emph{trigger axioms} that describe conditional event occurrences\\
- a set $\Gamma=\Gamma(0)\cup...$, denoting the \emph{observations}
at each timepoint, \\
- a set $\Delta_1=\Delta_1 (0)\cup...$, denoting the narrative of
action occurrences, and\\
- a set $\Omega$ of unique names axioms. \hfill
\end{definition}

Explanation closure axioms are created by means of circumscription
to minimize the extension of all $Initiates$, $Terminates$,
$Releases$, $Happens$ predicates. That is, the events and observations that are known at the time of reasoning are all that the agent assumes to have happened in the world; if new information is obtained in the future, the agent can revise its inferences about the world state.

A \emph{knowledge base} $KB(T)$ is a set of ground facts (i.e.,
fluents and events) and represents the state of the world at
timepoint $T$ and the events that are planned to occur at this
timepoint.

A fundamental extension of most action theories, vital for
real-world domains, is related to their ability to refer not only
to what an agent knows, but also to what it does not know \citep{LevesqueLake07Handbook}. This requires the
modeling of an agent's epistemic notions and an account of its changing state of knowledge by means of both ordinary
and knowledge-producing (sense) actions. The possible-worlds based model is a commonly used approach to represent epistemic notions in formal logics. An epistemic extension of the Event Calculus under the possible worlds semantics has recently been proposed by \citet{MillerLP13}.

Although highly expressive, the possible worlds model is generally computationally intensive \citep{PetrickL02}. 
The framework in this paper relies on the Discrete Event Calculus Knowledge Theory (DECKT) \citep{patkos09}, which develops
an epistemic extension of $\mathcal{DEC}$, using a deduction-oriented rather than
possible-worlds based model of knowledge. DECKT employs a meta-approach that
axiomatizes explicitly knowledge change rather than using possible world
semantics and modal operators.

\section{A Rule-Based Production System for the Event Calculus}
\label{sec3}

In this section, we describe Cerbere, a general-purpose rule-based reasoner for the Event
Calculus designed to perform causal, temporal and epistemic reasoning tasks with information obtained at run-time. Cerbere implements both the non-epistemic Event Calculus axiomatization and the DECKT epistemic extension. Within our logic-based forward-chaining framework, Event Calculus epistemic and
non-epistemic effect axioms, state constraints and domain rules are compiled
into production rules, preserving the rich semantics of the underlying formalisms. In contrast with ordinary rule-based systems deployed for
reactive reasoning in dynamic worlds, where the actions that lead to
the assertion and retraction of facts in a KB have no real semantics and
high-level structures, Cerbere uses the high-level
structures of $\mathcal{DEC}$ and DECKT to define the causal
properties of actions and events or to distinguish between ordinary
and epistemic facts that are initiated, terminated or triggered, in a context-dependent manner.

\begin{figure}[t]\begin{center}
\includegraphics[width=3.5in]{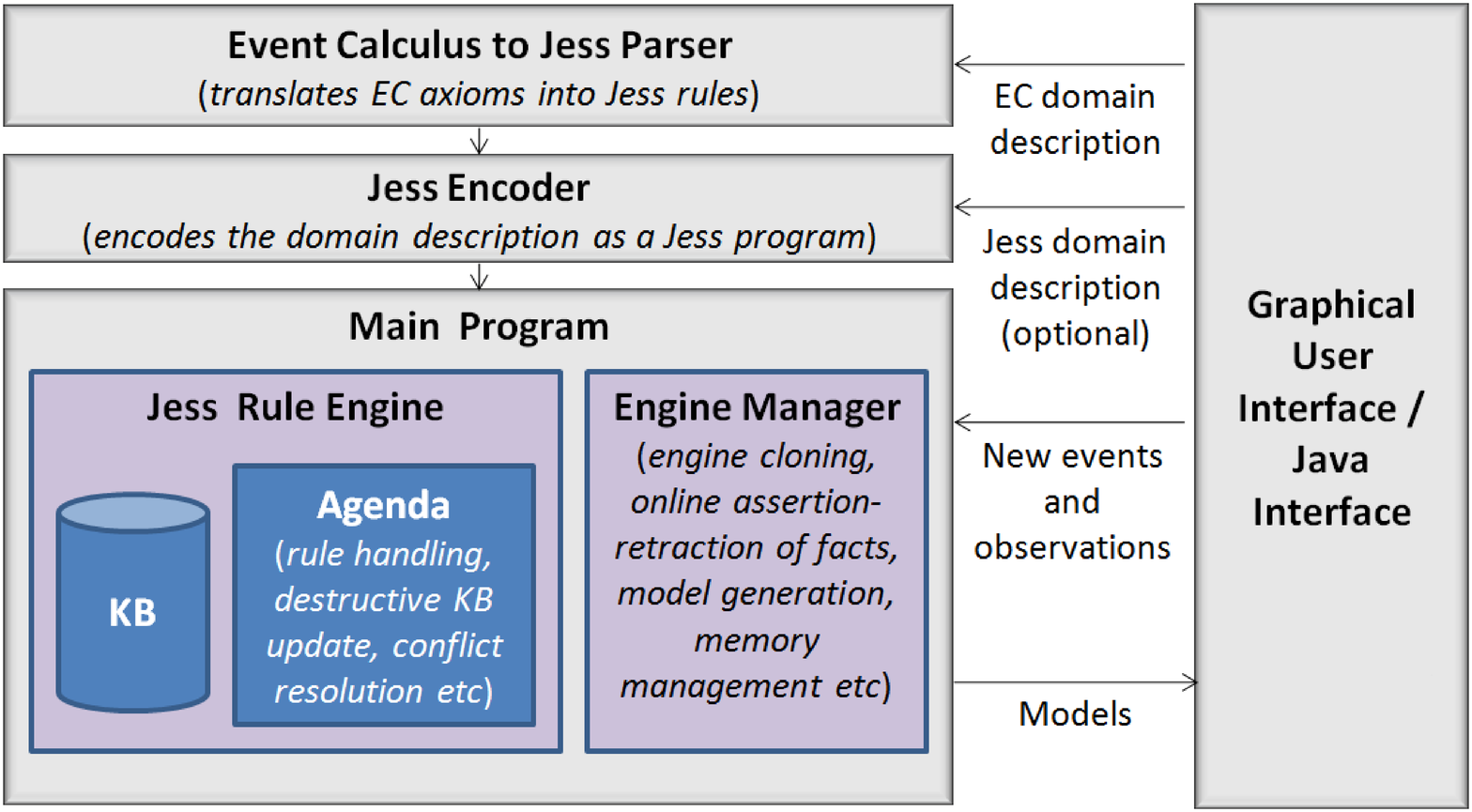} 
\caption{Cerbere architecture.} \label{reasonerArch}\end{center}
\end{figure}

Cerbere\footnote{\url{http://www.csd.uoc.gr/~patkos/TPLP15Appendix.htm} } comprises different modules to facilitate developers in
the construction of Event Calculus theories and in the monitoring of system execution at runtime. $\mathcal{DEC}$ and DECKT axiomatizations are implemented on top of
Jess\footnote{Jess: \url{http://www.jessrules.com/} (last accessed:
November 2015)}, an efficient rule engine that applies an enhanced version
of the Rete algorithm \citep{Forgy91} to process rules. Event Calculus programs are translated into Jess rules and encoded appropriately for execution by the underlying rule engine. As a consequence, the resulting programs inherit the
declarative epistemic and non-epistemic semantics, while the
reasoning cycle described next enables for online reasoning. Domain independent axioms of our formalisms are pre-compiled as Jess rules.

\subsection{The Reasoning Cycle}
\label{sec2OpSem}

Fig. \ref{reasonerArch} shows the main components that constitute the reasoner. The KB of Cerbere is structured as a deductive database. A rule
engine matches facts in the working memory, event narratives and
observations arriving on-the-fly with conditions of rules, deriving
the resulting world state and the events that are or may be
triggered. The reasoning cycle implements a model generator to construct
all possible models that satisfy the given narrative, the set of causal
and temporal constraints and the observations obtained at runtime. When multiple models are being inferred at a given timepoint, a new Jess instance with its own KB and agenda is created, coordinated by the Engine Manager. New actions and observations cause the update of the pool of valid models, in order to reflect the changes brought about.

{\tiny
\begin{figure}[t]\begin{center}
\algsetup{linenosize=\footnotesize}
  \footnotesize
\begin{algorithmic}[1]
\LOOP
\STATE obtain information from external sources at timepoint $t$ and update $\Gamma(t)$ and $\Delta_1(t)$ sets accordingly
\FORALL{models $KB(t)$ in the pool of valid models at timepoint $t$}
\IF {($KB(t) \cup \Gamma(t)$ is not consistent)}
\STATE remove $KB(t)$ from the pool of valid models
\ELSE
\STATE apply $\mathcal{DEC}$ axioms to $KB(t) \cup \Delta_1(t)
\cup \Sigma$, to obtain $KB^*(t+1)$
\STATE apply $\Psi$ to $KB^*(t+1)$ to obtain $KB^{**}(t+1)$
\STATE produce all combinations of the truth values for the remaining fluents and store them in the set $Comb$
\FORALL{elements $Cb$ of $Comb$}
\IF{$KB^{**}(t+1) \cup Cb$ is consistent with $\Psi$}
\STATE $KB(t+1) \leftarrow KB^{**}(t+1) \cup Cb$
\STATE store $KB(t+1)$ in the pool of valid models
\STATE apply $\Delta_2(t+1)$ trigger axioms to $KB(t+1)$ to produce events that are triggered at $t+1$ and append them in $\Delta_1(t+1)$
\ENDIF
\ENDFOR
\ENDIF
\ENDFOR
\ENDLOOP
\end{algorithmic}
\end{center}
\caption{Cerbere reasoning cycle.}
\label{opSem}
\end{figure}
}

The system's reasoning cycle controls the workflow of actions appropriate for run-time monitoring of a dynamic environment. Fig. \ref{opSem} shows the loop for the non-epistemic case. As a first step (line 2), information obtained from sensors, actuators, communication modules or other sources is added to the already planned actions to be performed, in order to be included in the reasoning process. The assimilation of observations regarding the state of fluents
requires careful treatment within an online system: $\mathcal{DEC}$ does not allow for inertial (i.e., not released) fluents to
modify their state unless an event explicitly interacts with them; consequently, for the non-epistemic component, where complete world
description must be preserved at all times, observations about
fluents that contradict stored derivations lead to model
elimination (lines 4-5). This technique is applied to resolve ambiguity caused by partial world descriptions or non-deterministic effects of actions.

Next, an incremental construction of the world state is implemented (lines 7-12). First, all inertial fluents at timepoint $t+1$ are determined and stored in $KB^*(t+1)$ (line 7), and then all indirect effects of actions are added, due to the state constraints that are triggered at $t$ (line 8). All combinations of the truth values of those fluents that are neither inertial nor subject to an activated state constraint of $\Psi$ at $t+1$ form the elements of the $Comb$ set (line 9). Finally, the models that are inconsistent with the domain state constraints are discarded (lines 10-13) and, for the remaining, triggered actions are considered (line 14).

Given a consistent domain theory, it is easy to show that the cycle terminates after each timepoint $t$ (the number of KBs produced at step 9 is limited by the number of fluents, no fluent is reused and there is no recursion). The same reasoning cycle is also applicable to epistemic reasoning, with the difference that DECKT meta-axioms are also accounted for at steps 7-14 and that observations at the initial step are not used for model elimination, but rather for knowledge update of those world aspects that are unknown at that timepoint. It should be noted that, according to the DECKT approach, unknown world fluents are reified as epistemic fluents. As a result, in the epistemic case a single KB is maintained at all times (i.e., $Comb$ in line 9 never contains more than one elements).

\subsection{Complexity Analysis}
\label{sec:complexity}

Our implementation supports the dialects of the Event Calculus discussed in Section \ref{background}, in order to perform temporal projection with events arriving at chronological order. Despite the nested exponential algorithm shown in Fig. \ref{opSem}, this situation only appears in the non-epistemic case for domains with high uncertainty, as we describe next; for the epistemic case, the performance is exponential at worst. Moreover, inferencing can still be computable in most domains of interest, as verified in our experimental evaluation in Section \ref{sec:experiments}, due to specialized features for handling fluent generation and destruction, along with the fact that all theories are reduced into propositional.

In more detail, steps 7 and 8, the main inference tasks in the reasoning cycle, carry out deductive closure (materialization) on ground facts to produce all possible derivations that make implicit information explicit. This is accomplished by the Jess inference mechanism, which relies on intensive use of hash tables and caching to reduce time complexity of pattern matching. In particular, the Rete algorithm implementation that is being used has linear computational complexity in the size of the working memory: according to the Jess manual,\footnote{Ernest Friedman-Hill, Jess - The Rule Engine for the Java Platform version Version 7.1p2,  \url{http://www.jessrules.com/jess/docs/Jess71p2.pdf}} finding which rules fire given a number of $F$ facts in the working memory is in the order of $O(RFP)$, where $R$ is the number of rules and $P$ the average number of patterns in the body of rules.\footnote{An attempt to map parameters $F,R,P$ of Jess with the actual Event Calculus axioms that Cerbere translates requires a detailed elaboration of the parsing process, which falls beyond the scope of this article. Nevertheless, Appendix A.5 gives the general picture of this parsing. The reader can notice there how axioms that involve nested preconditions along with negation and variable comparisons, result in rather complex rule formulations.}

From the implementation standpoint, it is worth noting that reasoning at each step operates in a monotonic space and no information that has been inferred for a
given timepoint is deleted at that timepoint. This enhances the behavior of the system,
considering that pattern-matching in Jess is performed
while changes are being made in the working memory, rather than just at
the actual execution time.

Given the materialized KBs of past timepoints, determining if new observations are consistent with the already inferred facts (step 4) becomes trivial, by querying whether their negation exists in the KB. This query does not need to search through the whole KB, since no inference is involved. Similarly, at step 7, a query is issued with each triggered state constraint to verify that no newly inferred fact contradicts itself in the already established KB.

As evidenced in Fig. \ref{opSem}, the predominant complexity factor for the non-epistemic case in general is the number of released fluents that cause the creation of multiple KBs at step 9. This aspect characterizes the degree of non-determinism regarding the effects of actions in a given domain: for a domain of $n$ fluents, an exponential number of KBs may be produced at each timepoint in the worse case, when all effects are non-deterministic. This is unusual to meet in practical systems. Moreover, Cerbere's architecture relies on the generation of a new Jess instance for each model that is inferred, as mentioned above. All these instances are independent from one another and controlled by the Engine Manager component (Fig. \ref{reasonerArch}), offering the possibility for extensive use of parallelization techniques for both data and computation distribution, a direction that we plan to exploit in the future.

For the epistemic case, on the other hand, a single KB is always preserved, yet an exponential number of epistemic fluents may need to be stored in the worst case, since disjunctions of domain fluents are treated as ordinary fluents (represented as \emph{hidden causal dependencies} \citep{patkos09}). These are created either due to the execution of actions with unknown
preconditions or by the sets of interrelated state constraints involving fluents whose truth value is unknown. From the practical perspective, the latter are expected to form groups of small sizes (so called \emph{dominos domains} 
which lead to chaotic environments are not commonly met in
commonsense domains). As we show in Sec \ref{sec:experiments}, the main performance factor of our reasoner is not the size of the domain but rather the number fluents that need to be updated, which leads to efficient execution even in large domains.

\subsection{System Added Features}

\begin{figure}[t]\begin{center}
\includegraphics[scale=0.27]{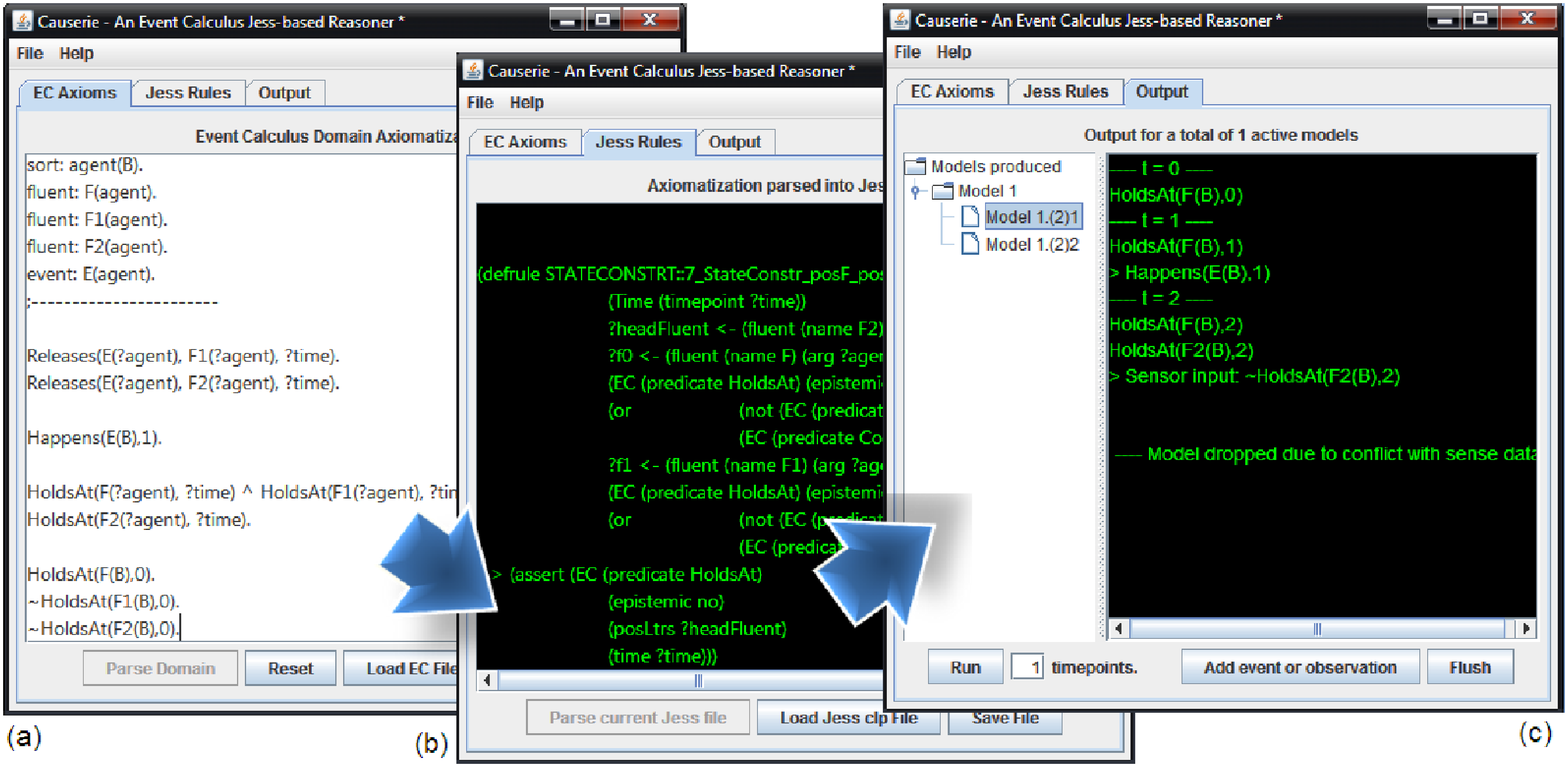} 
\caption{The user interface facilitates developers in writing EC
axioms, parsing them as Jess rules and offering online reasoning
functionalities.} \label{reasonerUIs}\end{center}
\end{figure}

A graphical user interface supports the developer during all steps of the design and implementation cycles. Fig. \ref{reasonerUIs} displays the visual development environment of Cerbere through a typical interaction loop: the developer designs a domain axiomatization by means of an intuitive Event
Calculus syntax (Fig. \ref{reasonerUIs}a), which the system parses into appropriate
Jess rules (Fig. \ref{reasonerUIs}b) and finally, at execution time, the programmer is
informed about the progress of reasoning and the elicitation of
commonsense knowledge (Fig. \ref{reasonerUIs}c).

The visual development
environment enables the programmer to implement the mental state of
rational agents at a more abstract level by parsing only
Event Calculus axiomatizations and, optionally, by modifying specific
rules of the Jess program for specialized tasks. New events and
observations can be asserted on-the-fly either manually, by interacting with
the user interface, or through a Java interface that enables the
system to automatically manipulate information arriving from sensors and actuators.
Before parsing, the user can choose either the execution of
classical Event Calculus reasoning or epistemic reasoning, i.e., to
incorporate the DECKT axiomatization or not.

A multitude of features have been integrated to the reasoner, in
order to enhance the implementation of its operational behavior,
while still maintaining consistency with the basic tenets of its
axiomatizations. For instance, proper memory handling is crucial for any online system that operates for long periods of time: overloading the memory with ground terms, as a result of inference making, can quickly deteriorate performance. Cerbere offers the ability to ``flush'' ground terms stored in the memory of the rule engine that are older than a specific timepoint. Combined with the semi-destructive update of the KB, described next, these techniques enhance the reasoner's performance. 

\subsubsection{Combining Reactive Reasoning with Multiple Models}


Most reasoning tools for logic programming are able to generate all valid models that are permitted by their semantics given a particular problem instance. ASP reasoners for example produce all answer sets that are valid for a specific program. This is not typically the case for rule-based reasoners, such as Jess or Drools, which are primary deployed to implement the reactive behavior of a system. Given an occurring set of events, rule triggering in these systems typically happens in a deterministic space and, thus, multiple alternative models are not generated.

Cerbere, being at the convergence of both logic- and rule-based reasoning, is designed to support multiple model generation. The use of released fluents, in particular, that introduce non-determinism in the truth value of fluents, necessitate the creation of alternative conclusions. By coupling possible model generation with the ability to perform sensing at execution time, the reasoner offers a powerful feature to online reasoning systems: models that contradict information arriving from external observations are automatically dropped, enabling reactive systems to handle uncertainty and adapt to knowledge arriving on-the-fly. Fig. \ref{reasonerUIs}(c) for instance, displays in the left column a tree structure where the user can select which of the valid models to display in the main window; the one shown in this particular snapshot will not be considered in future executions, as its conclusions at timepoint 2 contradicted the sensor input that arrived.

To accomplish this functionality, Cerbere creates and destroys different instances of the Jess rule engine at run-time, each representing a distinct valid model. The Engine Manager module (Fig. \ref{reasonerArch}) monitors the lifecycle of these engines that operate based on their own dedicated KB and rule-execution agenda.



\subsubsection{Semi-destructive KB Update}

The support for both epistemic and non-epistemic
derivations into a production framework, as well as the requirement
for run-time execution, led to the introduction of alternative
mechanisms to the reasoning process, which are not typically met in
the signature of the original formalisms. For instance, instead of
implementing parallel circumscription of predicates as employed by
standard Event Calculus, negation-as-failure (NaF) and the
semi-destructive update of the KB, which are encompassed in our
rule-based system, offer a solution to the computational frame
problem without the need to write additional frame axioms to perform
predicate completion. With application of NaF, all fluents that do not exist in the KB are regarded as false by default; only the fluents that hold at a particular time instant are stored as facts, thus reducing reasoning effort.

The semi-destructive update of the KB is an optional choice for the user, which can activate or de-activate it at run-time. This technique maintains a single snapshot of the world state, updating only the values of those fluents that are affected according to the narrative. This technique refers exclusively to inertial parameters (line 7 in Fig. \ref{opSem}) and is implemented following the next procedure, as introduced by \citet{KowalskiSadri10}:\footnote{The existence of state constraint in our theories calls for an extended definition in comparison to the one given by \citet{KowalskiSadri10}. As these constraints cannot be updated destructively, our notions led to a so-called `` semi-destructive'' approach.}
\begin{tabbing}
$KB^*(t+1) = $\\
$(KB$\=$^*(t) - \{$\=$f | Happens(e,t) \in \Delta_1(t),$\\
\>\>$Terminates(e,f,t) \vee Releases(e,f,t) \in \Sigma,$\`(3.1)\\
\>\>and all $f_i \in C_f$ hold in $KB(t)\}) \cup$\\
$(\{f | Happens(e,t) \in \Delta_1(t),
Initiates(e,f,t) \in \Sigma,$ and all $f_i \in C_f$ hold in
$KB(t)\}).$
\end{tabbing}

Notice that in this process only positive fluents are asserted in the knowledge base. Negative fluents are assumed to be false by
application of NaF, therefore they need not be explicitly
introduced. While the semi-destructive generation of successive world states is a very efficient choice in terms of memory storage and execution times (see Section \ref{sec:experiments}), it is not appropriate for all types of reasoning tasks, as for instance when we need to keep the history of conclusions. During parsing, the reasoner identifies whether there are axioms that refer to past timepoints and informs the developer that the semi-destructive option has been deactivated, to safeguard the correct execution of the axiomatization.

\subsubsection{Epistemic Reasoning}

\begin{figure}[ht]
\centering
\subfigure[Knowledge evolution within Shanahan's circuit.] {
\includegraphics[scale=0.24]{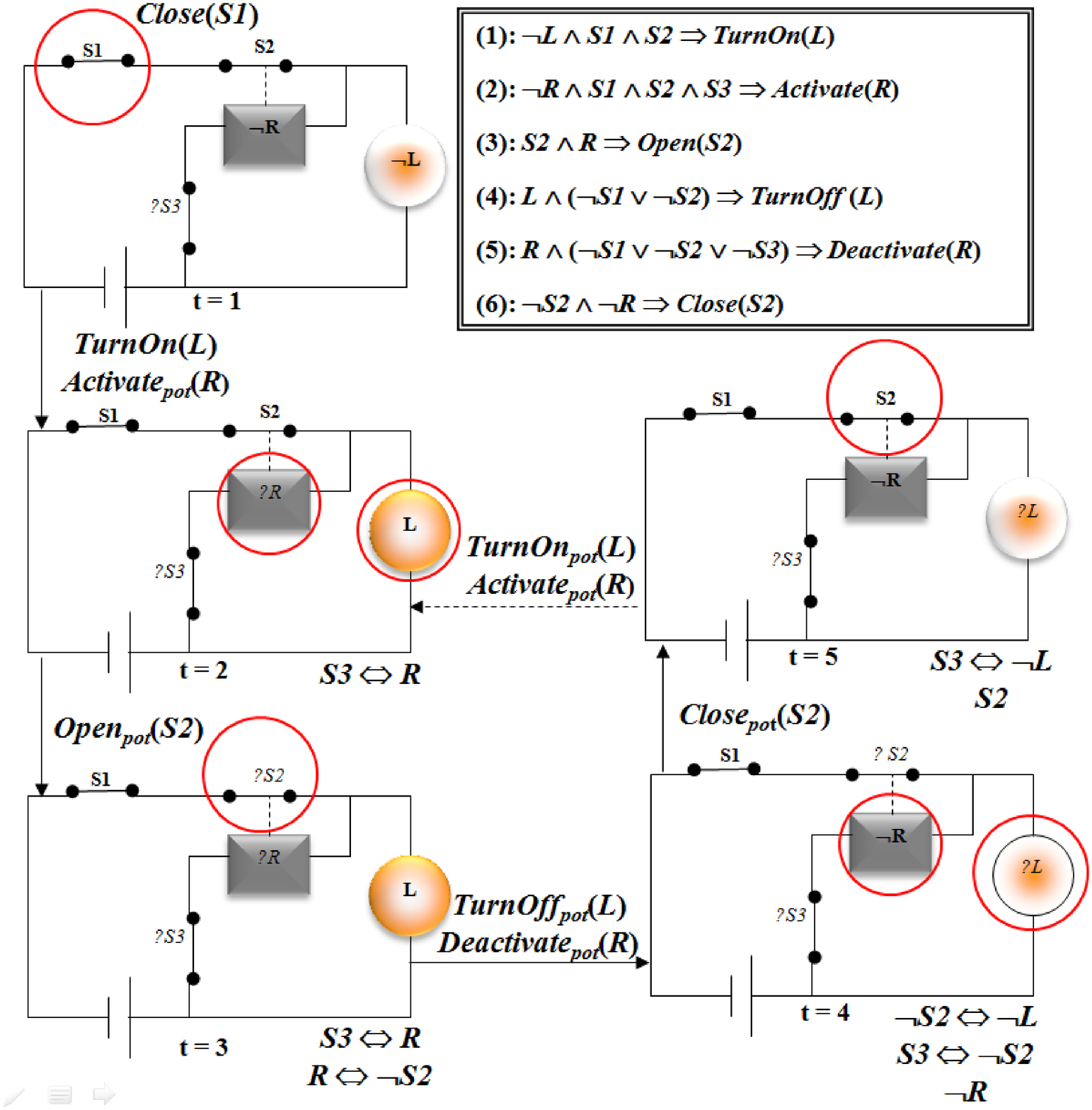}
 \label{shanahancirc}}
\quad
\subfigure[Cerbere output.]{
\includegraphics[scale=0.24]{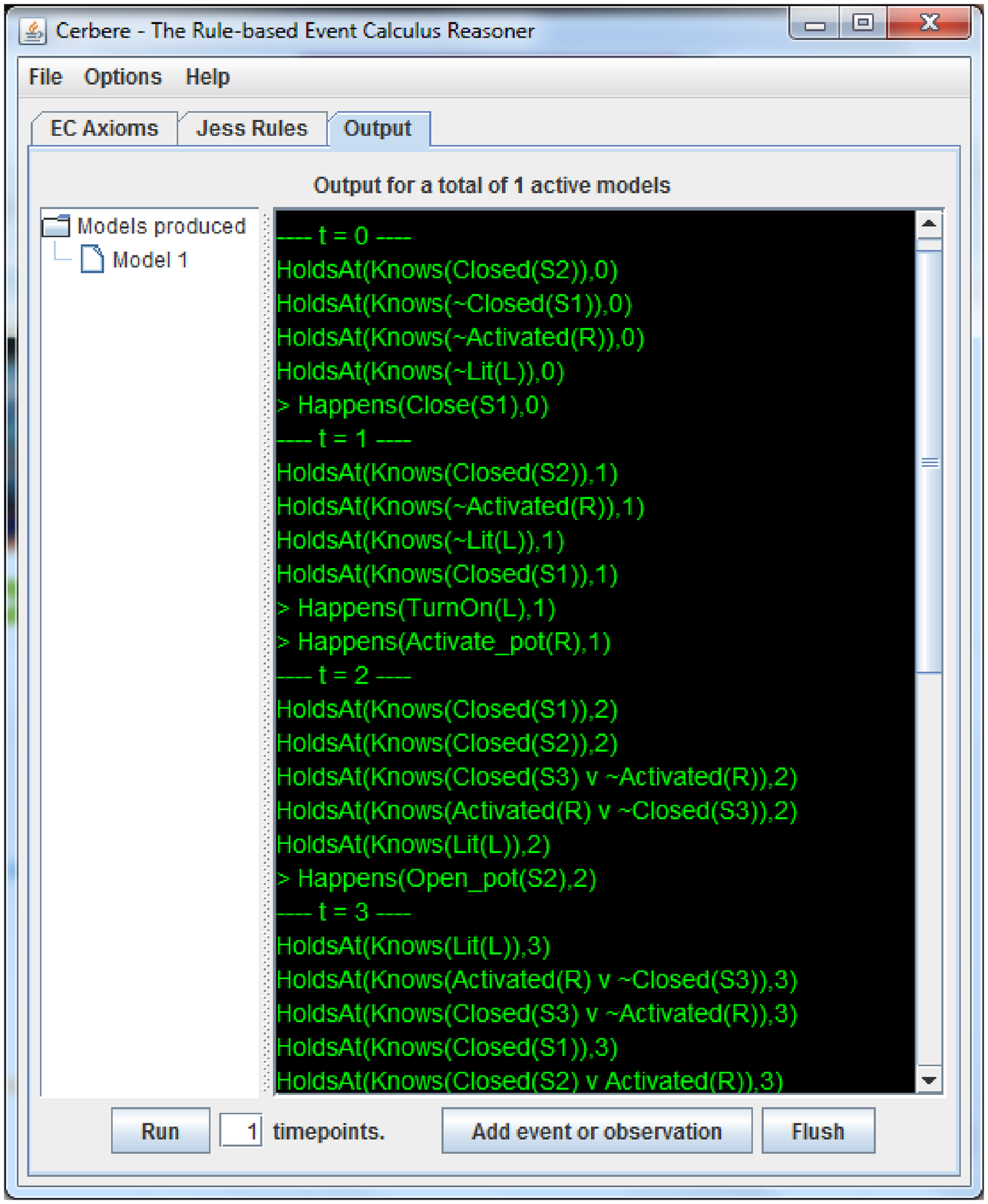}
\label{reasonerw} }
\caption{Shanahan's circuit with vicious cycles and delays.}
\end{figure}

To illustrate the capacity of Cerbere to perform epistemic reasoning, we axiomatized a benchmark problem devised to study complex ramifications, and extended it in the context of partial observability. The so-called Shanahan's circuit shown in Fig. \ref{shanahancirc}, is a variation of Thielsher's circuit \citep{ThielscherBook00}, which involves delayed effects and cyclic fluent dependency: if
initially switch $S1$ is open, but $S2$ and $S3$ closed, closing
$S1$ leads to cycling ramification effects, ought to relay $R$, that
causes light $L$ to repeatedly become lit and unlit every 2 time
points. 
\citet{shanahan1999rpe}, as well as Mueller (\citeyear{Mueller06}, p. 120), presented a
proper behavior for this circuit using Event Calculus trigger axioms
under complete world knowledge. Yet, a question raised in Shanahan's original paper about the possible inferences that can be made if the initial state of $S3$ is unknown, requires epistemic reasoning, in order to be answered.

Despite its seemingly simplistic configuration, the combination of
ramifications with partial observability introduces a
challenging commonsense problem: (a) delayed indirect effects call
for an explicit treatment of time, (b) lack of knowledge requires
for an epistemic theory to reason about potential event occurrences
and epistemic correlations among fluents, and (c) the instability of
the domain causes the correlations to be combined, potentially
producing explicit knowledge. To our knowledge, only the formalism developed by \citet{MillerMP13} may accommodate the involved phenomena, yet relying on a possible worlds-based approach.

Fig. \ref{shanahancirc} displays the conditionally stable cycling domain of
Shanahan's circuit at successive timepoints. Fluents that are
known are shown in bold, unknown fluents appear in italics with a
question mark, while circles mark fluents whose epistemic state has
changed at that
timepoint. The box on the upper right corner shows abbreviated the trigger
axioms that characterize the circuit's behavior.

The domain axiomatization comprises a number of effect axioms, e.g.,
\begin{tabbing}
$Initiates(Close(s), Closed(s),t)$), $Terminates(Open(s), Closed(s),t)$\`(3.2)
\end{tabbing}
and a number of trigger axioms, e.g.,
\begin{tabbing}
$\neg HoldsAt(Lit(L),t) \wedge HoldsAt(Closed(S1),t) \wedge HoldsAt(Closed(S2),t) \Rightarrow$\`(3.3)\\
$Happens(TurnOn(L),t)$
\end{tabbing}

Initially, the state of all fluents, except $S3$, is known to the
agent. The role of the epistemic theory is to determine which
trigger events \emph{are known} to occur, which \emph{may} occur, giving rise to potential actions (e.g., $Happens(Activate_{pot}(R),1)$),
and which of their effects are known to the agent. The figure
displays the epistemic state of the agent at successive timepoints;
specifically, it shows the occurring events, snapshots of the
circuit that are stored in the KB and the implication rules that capture temporal causal relations between fluents, e.g.,
sensing either of $S3$ or $R$ at timepoint 1 will provide definite
knowledge about the other.

The partially-observable variation of Shanahan's circuit can be axiomatized using the DECKT theory, while Cerbere offers the leverage needed to support reasoning with the complex features involved. The reification of logical formulae, for instance, within the
predicates of the calculus is required, in order to represent  $HoldsAt(Knows( Closed(S3) \vee \neg Activated(R) ),2)$, without possessing explicit knowledge about any the involved fluents individually. Fig. \ref{reasonerw} shows a snapshot of the program's execution.

\section{Reasoning in Situation-aware Dynamic Environments}

To illustrate the potency and usability of the proposed system, we describe in this section a hybrid framework for reasoning in smart spaces that integrates Cerbere. The application domain of smart environments concerns context-aware sensor-rich spaces that adapt and respond to users' preferences and needs. The envisioned applications materialize a long anticipated application objective for AI, and many of the subproblems that emerge can be addressed by AI methods. An essential step towards exhibiting commonsense behavior and providing meaningful assistance to the inhabitants of smart spaces is to automate the recognition and understanding of the users' current state of affairs, which may involve simple or complex activities.

A plethora of methodologies and algorithms investigate activity recognition \citep{YeDMcK12,ChenK11,Sadri11,AggarwalHAA11,Yang09}. On the one hand, so called \emph{data-driven approaches} adopt primarily a probabilistic and statistical view of information and widely rely on the enormous impact of machine learning techniques in real-world applications \citep{LuF09,SinglaCS10}. Their ability to learn from datasets and their capacity to model uncertainty are two of their distinctive characteristics.
\emph{Knowledge-based approaches}, on the other hand, model the rules of inference from first principles, rather than learned from raw data, and typically rely on formal specifications of their syntax and semantics, exploiting symbolic modeling and logic-based reasoning techniques. The expressive power, along with the capacity to verify the correctness properties of their axiomatizations, are key advantages of these methodologies. 


Data-driven methods are currently the mainstream choice to activity recognition; yet, many activities are characterized by constraints and relationships among context data that can neither be directly acquired from sensors nor can be derived through statistical reasoning alone \citep{RiboniB11Cosar}. Even trivial user actions, such as the process of making coffee, pre-assume a significant extent of commonsense and domain knowledge with respect to their causal effects and ramifications. In addition, their compositions, often referred to as \emph{situations}, have rich structural and temporal aspects, such as duration, frequency and subsumption relations. A seamless integration of data-driven with knowledge-based methodologies is essential for the materialization of smart spaces. Much of current research is working towards this end (e.g., \citep{RiboniB11Cosar,RoyGBBPTB11,HelaouiRNBS12,SkarlatidisPVA11}.

Moreover, activity and situation awareness only pose one step towards the implementation of intelligent environments. Reacting effectively on exceptional situations is equally important, requiring a coherent approach to inference, sensing and actuation, as pointed out by \citet{PecoraCirilloDUS12}; still, not many approaches achieve to offer an integrated solution. The multitude of phenomena that can be expressed with the Event Calculus constitute the formalism exceptionally capable of providing support for many of the issues investigated above. Already, researchers start to exploit the potential of the Event Calculus to express both causal and temporal properties of events \citep{ArtikisSP10}.

Next, we present a hybrid framework that relies on a combination of the Event Calculus-based reasoning with probabilistic inferencing, in order to deliberate on possible activities with quantifiable uncertainty, but also to predict users' future actions and make decisions as to how best to support them. We first describe the probabilistic component that is based on Bayesian Networks (BNs) and then show how it is coupled with Cerbere in a hybrid framework for managing smart spaces.

\subsection{Enhanced Bayesian Networks}
\label{BNs}

Our decision to use BNs is primarily driven by their modeling simplicity and their ability to refer to the causal relations of the underlying entities, which match nicely with the reasoning style of Cerbere. More complex probabilistic methodologies can be used in the future if needed. Still, the model presented next goes beyond serving as a simple proof-of-concept for demonstrating a smooth integration; as classical BNs are rather restrictive to model the rich semantics of a real-world domain, we proceed by developing a variant of BNs, which we call \emph{enhanced} BNs (eBNs), that aims to promote expressiveness along with scalability and reusability.

Apart from estimating probability values for the nodes of a Bayesian graph, eBNs categorize nodes into different classes, in order to impose constraints of different type on them. We define a structure specifying how eBNs can be designed and composed, in order to form a network with the capacity to both \emph{recognize} and \emph{monitor} user activities in smart spaces. This design differentiates our work from other studies that apply BNs in a rather straightforward manner.

\begin{figure}[!t]
\centering
\includegraphics[width=3.5in]{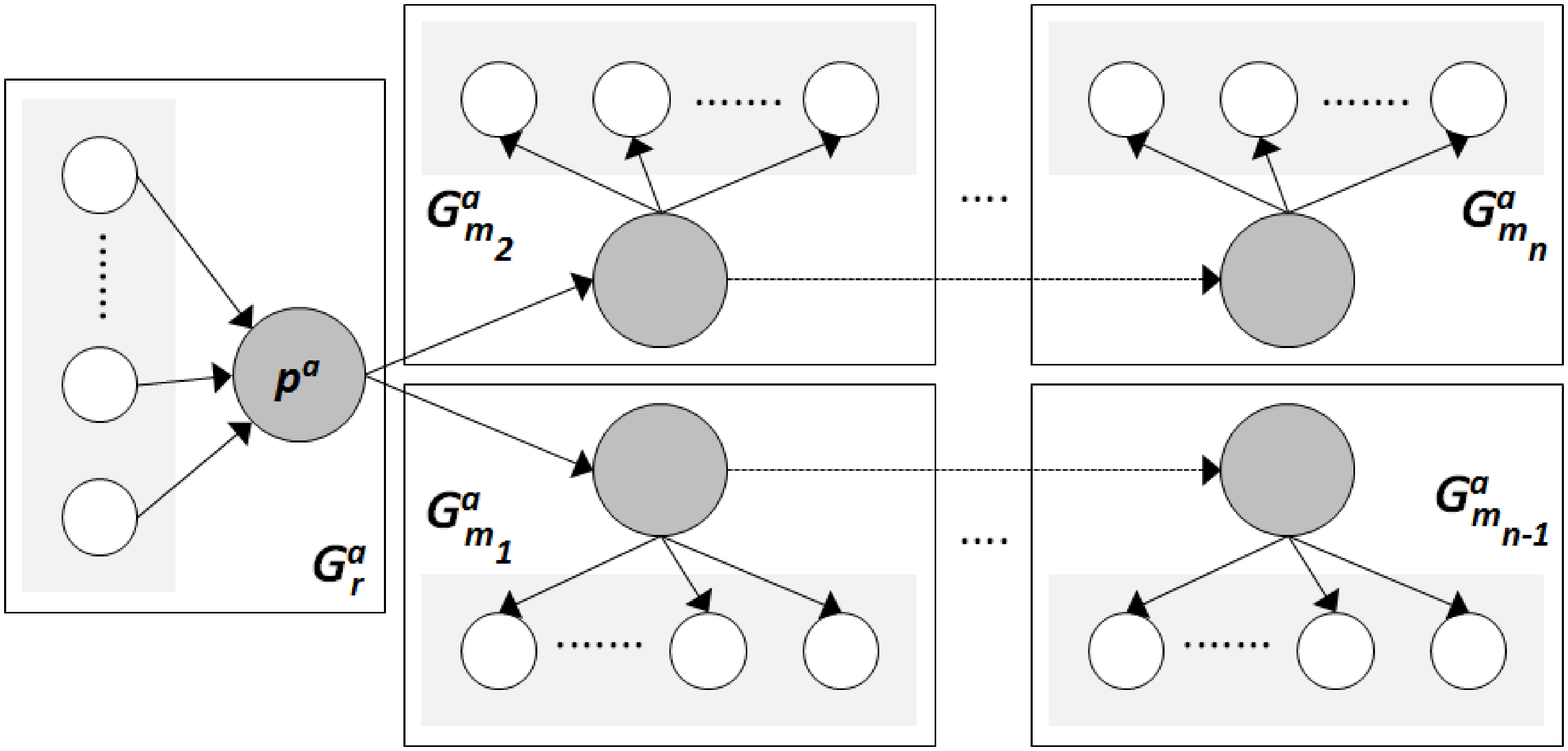}
\caption{Abstract layout of an Activity Network of eBNs for activity $a$, where $G^a_r$ is used for recognition of $a$, and $G^a_{m_i}$ for monitoring the progress of $a$.}
\label{abstractBN}
\end{figure}

In our modeling, the basic structure used to represent an activity is a composition of different eBNs, which we call Activity Network (AN) (Fig. \ref{abstractBN}). It comprises a set of eBNs: one of them is used for the recognition of some activity $a$ (graph $G^a_r$), while the rest are applied once the activity has been recognized with a certain confidence and contribute in calculating the most probable future actions the user may make at each step while performing the activity (graphs $G^a_{m_i}$). For example, the general activity of preparing breakfast can be composed of three phases: the first one, the preparation, is used for recognizing with certainty that the user is involved in this activity and, once this knowledge is established, the remaining phases, e.g., eating breakfast and arranging kitchenware into their places, are used to monitor the user, in order to assist by suggesting actions if needed.

In more detail, we model an AN $G^a$ for activity $a$ as a collection of eBNs: each $G^a$ has exactly one eBN $G^a_r$ for recognition purposes and zero or more eBNs $G^a_{m_i}$ for monitoring purposes, so that $G^a = \{G^a_r,G^a_{m_i}\}$ for $i\geq 0$. We assume that each node has a unique label, which represents an atomic sentence $p_i$, where an uppercase $P_i$ is used to denote either $p_i$ (when the sentence is true and satisfies all related to the node constraints) or $\neg p_i$ (when the sentence is false or some of the associated constraints is not satisfied).\footnote{For convenience, we use $p_i$ to represent both the logical sentence and the node, interchangeably.} Each eBN may consist of 4 disjoint sets of nodes: $V_f^a$, denoting state fluents (e.g., ``currently the time of day is morning''); $V_{act}^a$, denoting user activities (e.g., ``brushing teeth''); $V_e^a$, denoting actions (e.g., the action of opening the fridge door); and $V_g^a$, denoting groupings, i.e., typical patterns of the previous concepts (e.g., the opening and closing of the fridge door).

The assignment of nodes into distinct sets enables us to impose different types of constraints to the members of each set, which must be satisfied by the corresponding logical sentence, in order for the latter to be considered true. For instance, we can specify how many times an action has occurred before considering an action node to be satisfied (``moreThanXTimes'', ``lessThanXTimes'' operators), we can specify whether we want a fluent to hold or not to hold in order for the state fluent node to be satisfied, we can specify the duration and the interval within which an activity has or has not occurred for an activity node to be satisfied (``inTheLastXSec'', ``forAtLeastXSec'' operators), and others (more complex temporal and logical operators can be integrated in the future).

\begin{figure}[t]\begin{center}
\includegraphics[scale=0.34]{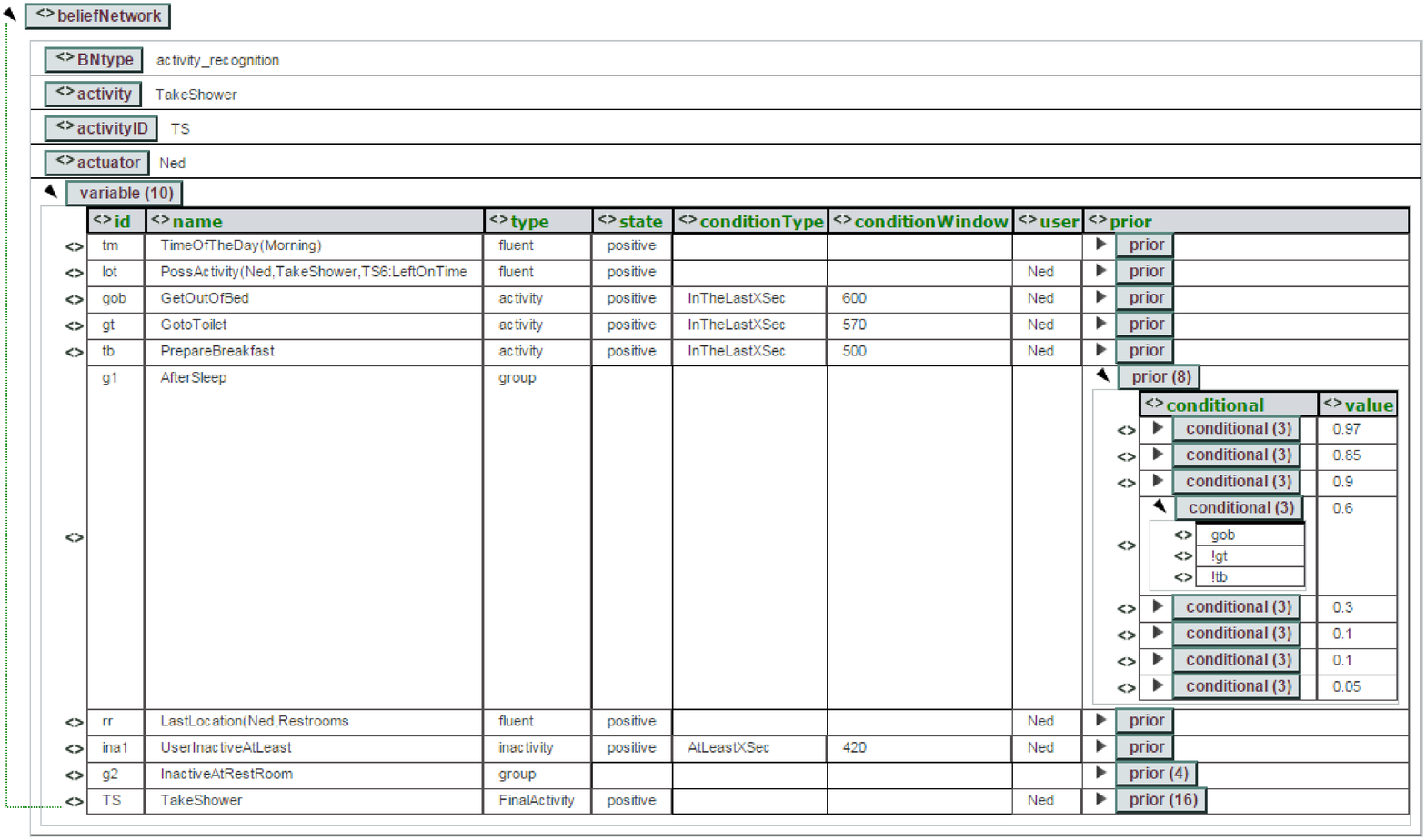} 
\caption{The tree representation of the XML file capturing a recognition BN (visualized with XmlGrid.net)} \label{XML}\end{center}
\end{figure}

By combining the different node types with the constraint operators, complex situations can be grouped, that can also be reused for implementing different eBNs or ANs. For example, we can group as a typical pattern the fact that when it is working day the user gets out of bed, goes to toilet and takes breakfast within 10 minutes, and takes shower after having breakfast, not before. The different BNs are modeled as XML files in our system; Fig. \ref{XML} shows a visualization of the XML file capturing a recognition BN for the $TakeShower$ activity, where the probabilities have been obtained through a training phase. Specifically, each node $p_i$ is characterized by the set of conditional probabilities: in Fig. \ref{XML}, $Pr(g1|gob \wedge \neg gt, \wedge \neg tb)=$ 0.6 meaning that the grouping node $g1$ mentioned before has a $60\%$ probability to be true if the user is out of bed (entry $gob$), but has neither gone to the toilet (entry $!gt$) nor had taken breakfast yet (entry $!tb$).

\begin{figure}[!t]
\centering
\includegraphics[width=3.5in]{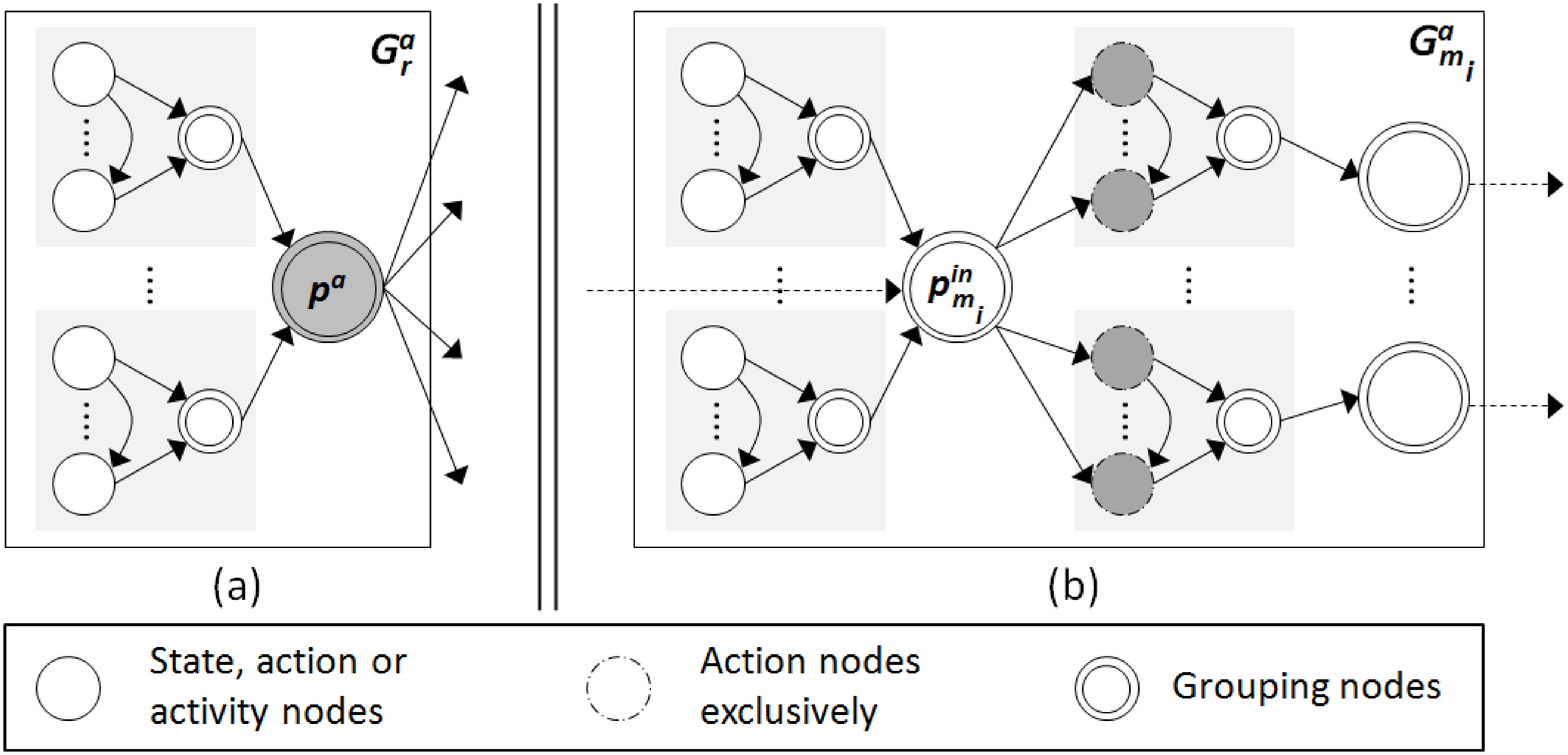}
\caption{(a) A recognition eBN, (b) A monitoring eBN.}
\label{BNtypes}
\end{figure}

\textbf{Activity Recognition:} One of our objectives is to use the eBNs, populated with the prior probabilities, in order to calculate the probability that an activity is taking place, given a set of observations at a specific time instant. Among the activity nodes $p \in V_{act}^a$ of $G^a_r$, there is exactly one node, labeled $p^a$, which depicts the activity $a$ of the $G^a$ and whose conditional probability we wish to calculate (Fig. \ref{BNtypes}(a)). Every $p \in V_g^a$ has $p^a$ as its only child, i.e., we only allow a single layer of grouping nodes, which has been found to be sufficient for modeling the majority of activities in a home setting. 

Given a set of (true or false) propositions $O = \{p^o_1,...,p^o_n\}$ that form the vector of observations at a particular time instant, such that $O \subseteq V_f^a \cup V_e^a \cup V_{act}^a \setminus p^a$, the objective of $G^a_r$ is to calculate the conditional probability $Pr(p^a|O)$. For those $p_j \not \in O$ for which we have no information about, we need to consider both their probability of being true and that of being false; to differentiate them from observations, we use capital letters to represent them, i.e., $P_j$ instead of $p_j$.

According to the Bayesian rule, the probability can be calculated as follows:

\begin{tabbing}
$Pr(p^a|O) = \frac{\sum J(\langle p^a, p^o_1,...,p^o_n,P_1,...,P_m\rangle)}{\sum J(\langle P^a, p^o_1,...,p^o_n,P_1,...,P_m\rangle)}$ \` (4.1)
\end{tabbing}

where $J(\langle P_1,...,P_k\rangle)$ is the joint probability distribution obtained directly from $G^a_r$ as follows:

\begin{tabbing}
$J(\langle p_1,...,p_k\rangle) = Pr(p_1|parents(p_1))\times ... \times Pr(p_k|parents(p_k))$ \` (4.2)
\end{tabbing}

The sum in the numerator and denominator of Eq. (4.1) refers to all possible probability distributions that result due to the combinations of the different $P_j$ that are unknown, i.e., those aspects that we have no observations for.

For instance, assume that for the $TakeShower$ activity (node $tsh$) we only have a single grouping node $g1$ with nodes $gob, gt, tb$ as its parents, as described before. Assume also that we only obtained observations for $gob$ and $\neg tb$ at a specific time instant. Then, $Pr(tsh|gob \wedge \neg tb) = \frac{\sum J(\langle tsh, gob, GT, \neg tb, G1\rangle)}{\sum J(\langle TSH, gob, GT, \neg tb, G1\rangle)}$, where
\begin{tabbing}
$J(\langle tsh, gob, gt, \neg tb, g1\rangle) = $\=$ Pr(tsh|g1)*Pr(gob)*Pr(gt)*(1-Pr(tb))*$\\ \>$Pr(g1|gob\wedge gt \wedge \neg tb)$\` (4.3)\\
$J(\langle tsh, gob, gt, \neg tb, \neg g1\rangle) = $\=$Pr(tsh|\neg g1)*Pr(gob)*Pr(gt)*(1-Pr(tb))*$\\ \>$(1-Pr(g1|gob\wedge gt \wedge \neg tb))$\` (4.4)\\
$J(\langle tsh, gob, \neg gt, \neg tb, g1\rangle) = $\=$ Pr(tsh|g1)*Pr(gob)*(1-Pr(gt))*(1-Pr(\neg tb))*$\\ \>$Pr(g1|gob\wedge \neg gt \wedge \neg tb)$\` (4.5)\\
$J(\langle tsh, gob, \neg gt, \neg tb, \neg g1\rangle) = $\=$Pr(tsh|\neg g1)*Pr(gob)*(1-Pr(gt))*(1-Pr(tb))*$\\ \>$(1-Pr(g1|gob\wedge \neg gt \wedge \neg tb))$\` (4.6)
\end{tabbing}
and similarly for the $\neg tsh$ case. At the end, $Pr(tsh|gob \wedge \neg tb)$ is the sum of the four joint probabilities listed above divided by the sum of all 8 elements that will result when $\neg tsh$ is also considered. A more elaborate illustration of this methodology is given in \citep{Brachman04}, chapter 12.

\textbf{Activity Monitoring: } The purpose of the $G^a_{m_i}$ graphs (Fig. \ref{BNtypes}(b)), on the other hand, is to calculate the probabilities of the actions the user has not performed yet. That is, given a set of observations $O$, each $G^a_{m_i}$ can be used to return $Pr(p_j|O)$, for each $p_j \in V^a_e \setminus O$. These graphs have an entry node $p_{m_i}^{in} \in V_g^a$, which denotes the start of the particular phase of the activity and zero or more exit nodes that connect the current phase with others. As a result, we can model both sequential execution of phases through the course of an activity (e.g., first preparing the table and then eating), as well as parallel or alternative phases (e.g., after eating the user can either wash the dishes or place them in the washing machine).

Similar to \citet{LuF09}, the proposed design of multiple ANs representing different activities does not enforce mutual exclusivity, thus does not preclude the monitoring of activities that may happen concurrently, such as preparing breakfast while talking on the phone. Moreover, the different eBNs are stored in the system's repository and can be reused, extended or refined based on the available training data without requiring reconfiguration of the network as a whole; new activities can be added with minimal changes to the existing design.

\subsection{A Hybrid Framework}

\begin{figure}[t]\begin{center}
\includegraphics[scale=0.27]{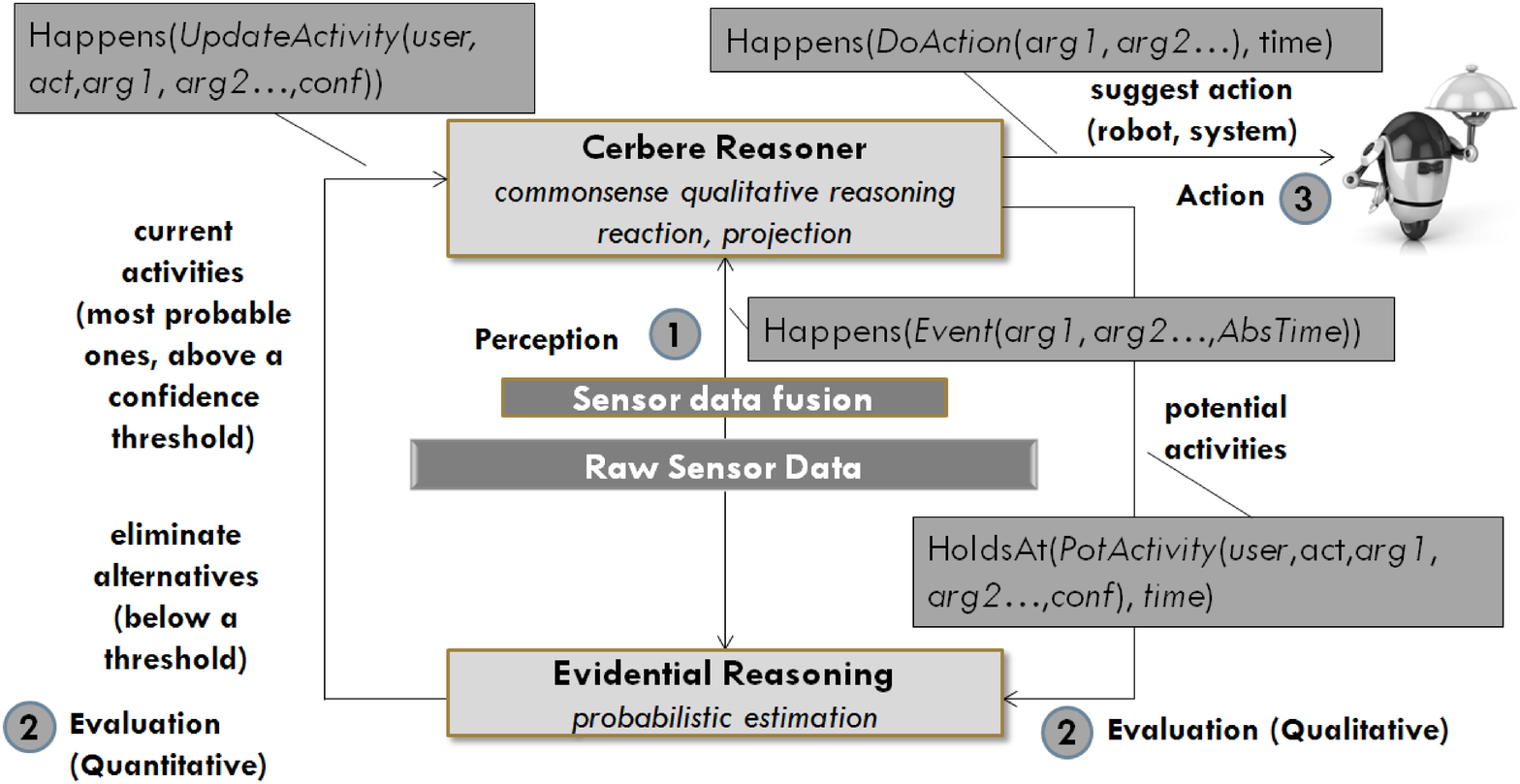} 
\caption{Information flow} \label{infoflow}\end{center}
\end{figure}

The coupling of Cerbere with the probabilistic component described above is intended to enable activity recognition, monitoring and actuation within smart spaces in a coherent cycle. This coupling progresses in three successive steps, as shown in Fig. \ref{infoflow}. The cycle is triggered by occurring events, as obtained by sensors that listen to various environmental parameters (even the lack of readings for a given amount of time may trigger appropriate events). Cerbere is assigned the task to evaluate various causal, spatial and temporal aspects, in order to determine which activities are possible at a particular time. First, a set of hard constraints is considered for each activity and those that satisfy all of them are evaluated through a set of soft constraints. At the end, the reasoner generates a list of $PossActivity$ statements for each possible activity. This feature aims at promoting flexibility in our system, as knowledge engineers can add or remove different rulesets of soft constraints without having to redesign the whole rule base for each activity. Even contradictory conclusions on activities are allowed, as they will be evaluated by the probabilistic component that will finalize the confidence estimation about them.

The logical derivations are annotated with an explanation and a reference value, taking the form $PossActivity(user, activity, explanation, weight)$. More specifically:
\begin{itemize}
\item{an ID is attached to the conclusions drawn by the reasoner that points to a predefined explanation justifying why the given conclusion has been reached. Formal theories are well known for their ability to justify their behavior and explain their derivations. A proper explanation can be used for building trust between the human user and the smart space, especially when the objective of the system is to persuade humans to perform certain tasks for their own benefit. Argumentation techniques can further be built on top of this information, as in similar studies \citep{MunozBA10,BikakisA10TKDE}.}
\item{along with the explanation, a weighted reference value is also given to rate the degree of how convincing the justification is for the given activity to take place. This weight ranges from 1 to 5 in our case, with a value of 1 denoting the highest degree of reliance.}
\end{itemize}

The platform then relies on probabilistic algorithms described in the previous subsection to calculate confidence values about these activities. Consider, for instance, the case where the user enters the bathroom in the morning after getting up from the bed. The following information will be produced as input to the probabilistic component, capturing the fact that the user has a preference for taking shower in the morning (value 2), no particular preference for brushing teeth at this time of the day (value 3), and is probable to be taking shower considering that he has not showered yet at that day (value 2):
\begin{tabbing}
$\models _{\mathcal{DEC}} $\=$HoldsAt(PossActivity(N$\=$ed,TakeShower,TS2$:$Morning,2),12) \wedge$\\
\>$HoldsAt(PossActivity(Ned,TakeShower,TS8$:$NoShowerYet,2),12)\wedge$ \`(4.3)\\
\>$HoldsAt(PossActivity(Ned,BrushTeeth,BT3$:$Morning,3),12)$
\end{tabbing}

With this information available, the reasoner provides a rich set of evidences to the probabilistic component to calculate the confidence values of the involved activities, while at the same time it manages to significantly narrow down the space of activities that need to be considered. It is worth noting that this aspect of our approach comes in contrast with the approach of \citet{RiboniB11Cosar}, where first a statistical prediction of \emph{all} activities is calculated and then ontological reasoning is used to eliminate counterintuitive or logically unfeasible results.

The activities that are above a threshold at the end of step 2 are considered accurately recognized. This set is returned to the reasoner to update its conclusions and proceed to a final step of inferencing, that of determining appropriate actions to perform ("DoActions" in Fig. \ref{infoflow}), in order to support the recognized activities. These may involve notifications or alerts directed to the user or changes in the state of devices that exist in the smart spaces.

\section{Experimental Results}
\label{sec:experiments}

We conducted a series of experiments to measure the domain-independent behavior of the reasoner in executing Event Calculus theories and also to evaluate its domain-dependent performance for accommodating typical reasoning tasks in a smart space under the proposed hybrid framework. We used an ordinary computer for our measurements, equipped with an Intel i5-2520M 2.50GHz CPU with 4Gb RAM.
\subsection{Domain-Independent Evaluation}

\textbf{Goal:} Our first concern is to study how the reasoner scales with the size of the domain, i.e., with the number of events, objects, fluents and axioms. The objective is to understand the behavior of the system and identify those features that affect performance, in general.

\noindent \textbf{Setting:} We measured the execution time in two similar settings. The first involved a single fluent type $F1$ and a single event type $E1$, but an increasing number of objects $O_n$. In a sense, this setting can resemble a space with sensors of the same type scattered in the environment, e.g., multiple location sensors. One event type modifies the truth value of fluents, i.e., $Initiates(E1(o),F1(o),t)$, where variable $o$ takes values from the set of $O_n$.

The second setting has a single object $O1$, but multiple fluent types $F_n$ and consequently different event types $E_n$ that affect each fluent respectively, leading to multiple effect axioms $Initiates(E_n(O1),F_n(O1),t)$. This setting can resemble a space equipped with completely different types of sensors measuring various features of the environment.\footnote{The axiomatization for both settings is available in the Appendix.}

\begin{table}[!t]\begin{center}\caption{Performance measurements when increasing the number of domain objects.}
\includegraphics[width=5in]{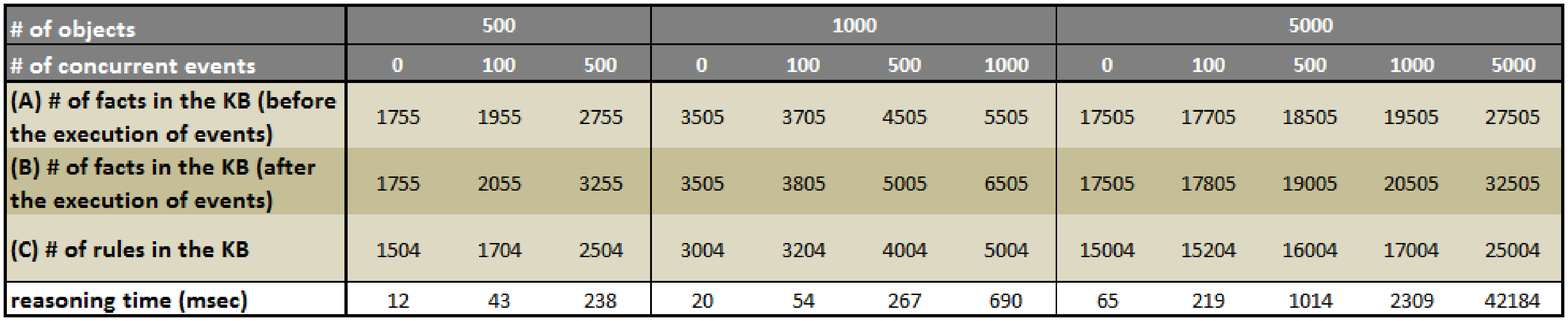} 
\label{tableOBJ}\end{center}
\end{table}

\begin{table}[!t]\begin{center}\caption{Performance measurements when increasing the number of domain fluent and event types.}
\includegraphics[width=5in]{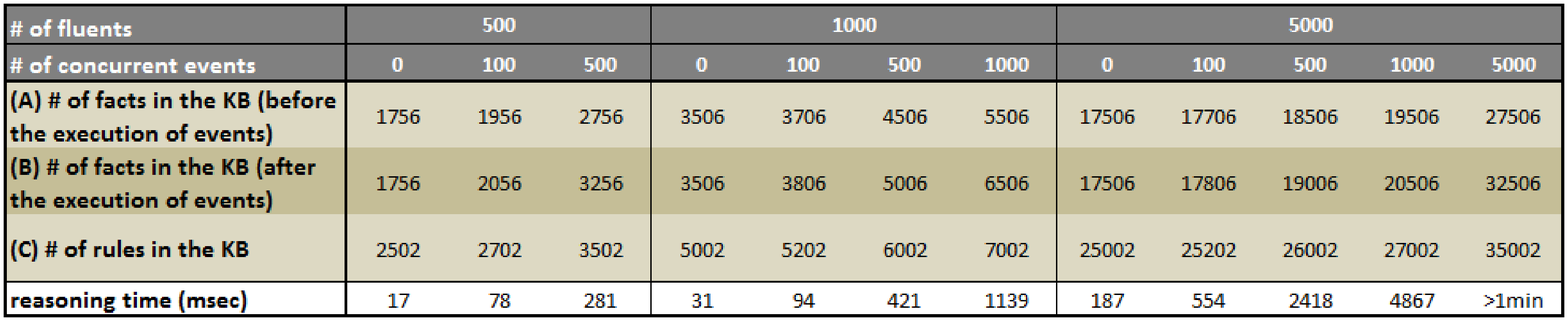} 
\label{tableFL}\end{center}
\end{table}

In both settings, $n$ ranges between 500, 1000 and 5000, as shown in Tables \ref{tableOBJ} and \ref{tableFL}. Moreover, we parameterized the number of concurrent events that the reasoner had to process at each timepoint, in order to update its KB. For instance, in Table \ref{tableOBJ}, the first 3-column set measures reasoning time in a domain of 500 objects, when no object changes its value (first column), when 100 objects change their value concurrently (second column), and when all 500 objects change their value concurrently (third column).

Because Cerbere stores in the KB only fluents that are true at a given timepoint, we split all occurring events in two sets of equal size, so that half of the events initiate their respective effect and the other half terminate it in order to obtain reliable results. For the same reason, the state of the KB before the occurrence of the events involves only fluents that may be terminated. For instance, the KB for the last 5-column set in Table \ref{tableFL} contains 2500 fluents before the action; the rest are false at that timepoint (the number of facts shown in row (A) results from various aspects, such as fluents, fluent definitions, events, event definitions, effect axioms, events about to occur and others, as explained in Appendix A.5). Finally, the semi-destructive update of the KB has been used, in order to have a clean starting point before the action (only the needed fluents and no other past data are stored).

\noindent \textbf{Conclusions:} There are two main conclusions that can be drawn from these experiments. First, updating objects is quicker than updating fluents (Table \ref{tableOBJ} vs Table \ref{tableFL}). This can be explained if we consider that in the former case a single effect axiom is triggered multiple times (for the different objects), whereas in the latter case the rule engine spends time trying to locate the effect axioms to trigger in its rule agenda.

Second, the reasoning time does not depend that much on the size of the domain, but rather on the number of changes that need to be made from one timepoint to the next. For instance, updating 100 objects concurrently, even in a domain involving a huge number of objects is less time consuming in comparison to updating 500 objects in significantly smaller domains (Table \ref{tableOBJ}). This is a desirable feature as it signifies the applicability of the reasoner in domains of realistic size. For instance, typical smart environments operate with a few hundred sensors, whose data is being preprocessed and fused before being fed to the logic-based component. This way, the chance for concurrent updates is reduced: the readings obtained by a location sensor are only relevant if a movement is detected and not during still periods. 

\subsection{Evaluation for Reasoning in Smart Spaces}

\textbf{Goal:} In this subsection, we assess the suitability of the proposed framework under the 3-step reasoning cycle to accommodate the requirements of smart environments. In such cases, activities are recognized by taking into consideration sequences of user actions, as well as past activity inferences. As such, our main concern now is response times when expressing sufficiently complex phenomena encountered in smart spaces. The accuracy of the inferred conclusions is not subject to investigation, since the coupling of symbolic reasoning with probabilistic estimations can enable the derivation of satisfactory conclusions by relying more on the latter when a rich set of training data is available or by engineering more detailed logic rules to model the desirable activities, otherwise.

\noindent \textbf{Setting:} We adopt the popular benchmark setup presented by \citet{vanKasteren08}, where the everyday activities of a young male human were recorded in a home setting for a period of almost one month. The generated dataset is based on information obtained by a number of binary sensors providing data about the state of doors, devices etc. and whose high-level interpretations were manually annotated by the user himself. To make the domain more realistic and closer to our targeted objectives, our examples are fabricated, in the sense that additional features are assumed without changing the domain. For example, while no abnormal or critical situations emerged in the course of the van Kasteren measurements, they are instead added in our rulesets to illustrate the ability of our system to exhibit proactive and/or assistive behavior.

We conducted our measurements given a typical narrative of actions. That is, instead of measuring the reasoning time after a random user action, which would not be illustrative since different actions trigger totally different number of rules with varying complexity, we recorded the reasoning time required after a narrative of 6 actual user actions, which we call a \emph{round} of actions. Each user action within the round initiates the 3-step cycle described before, causing many system-generated events to occur.

\begin{figure}[ht]
\centering
\subfigure[Sequential execution of activities.] {
\includegraphics[scale=0.33]{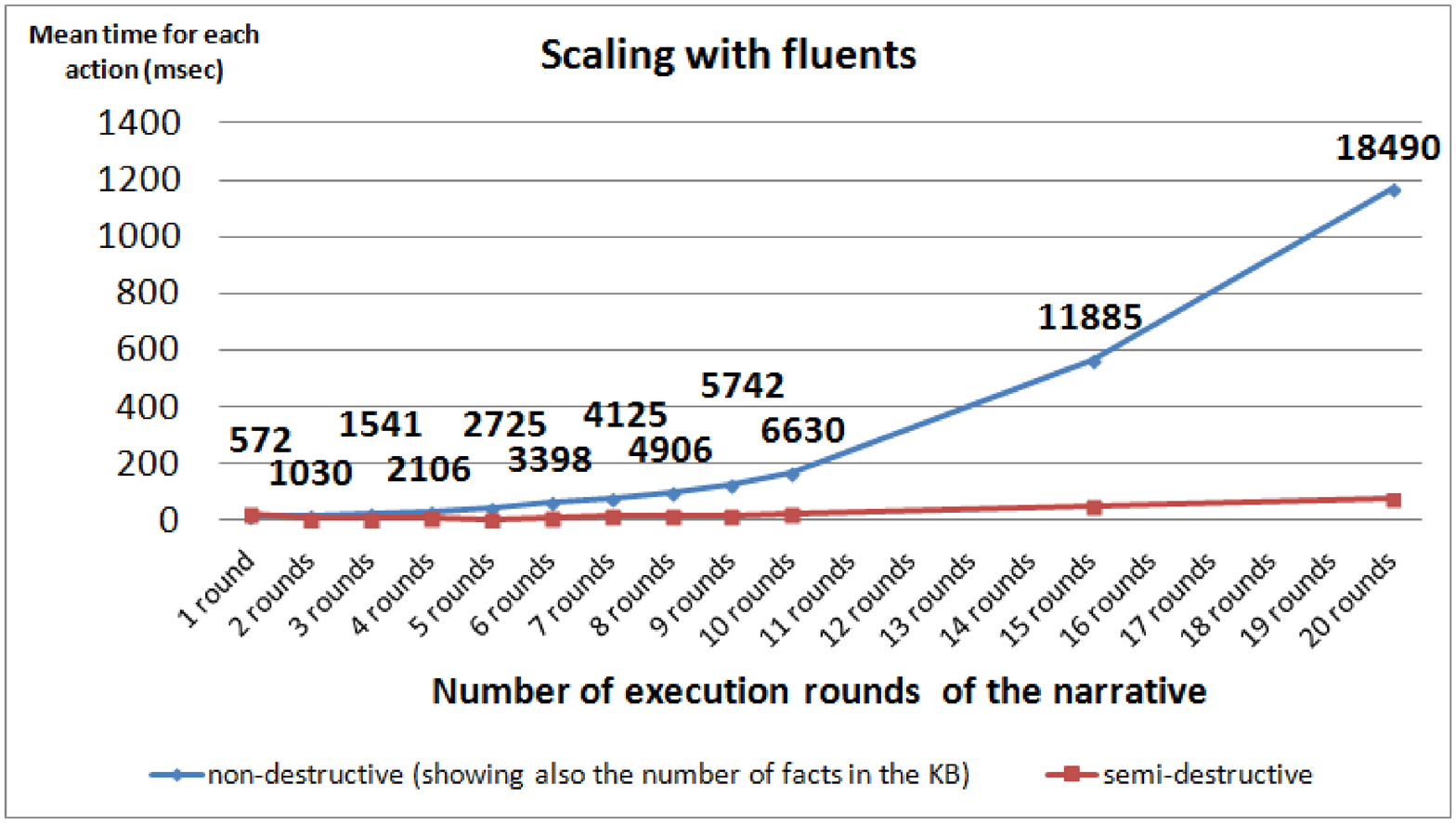}
 \label{graph1}}
\quad
\subfigure[Concurrent execution of activities.]{
\includegraphics[scale=0.35]{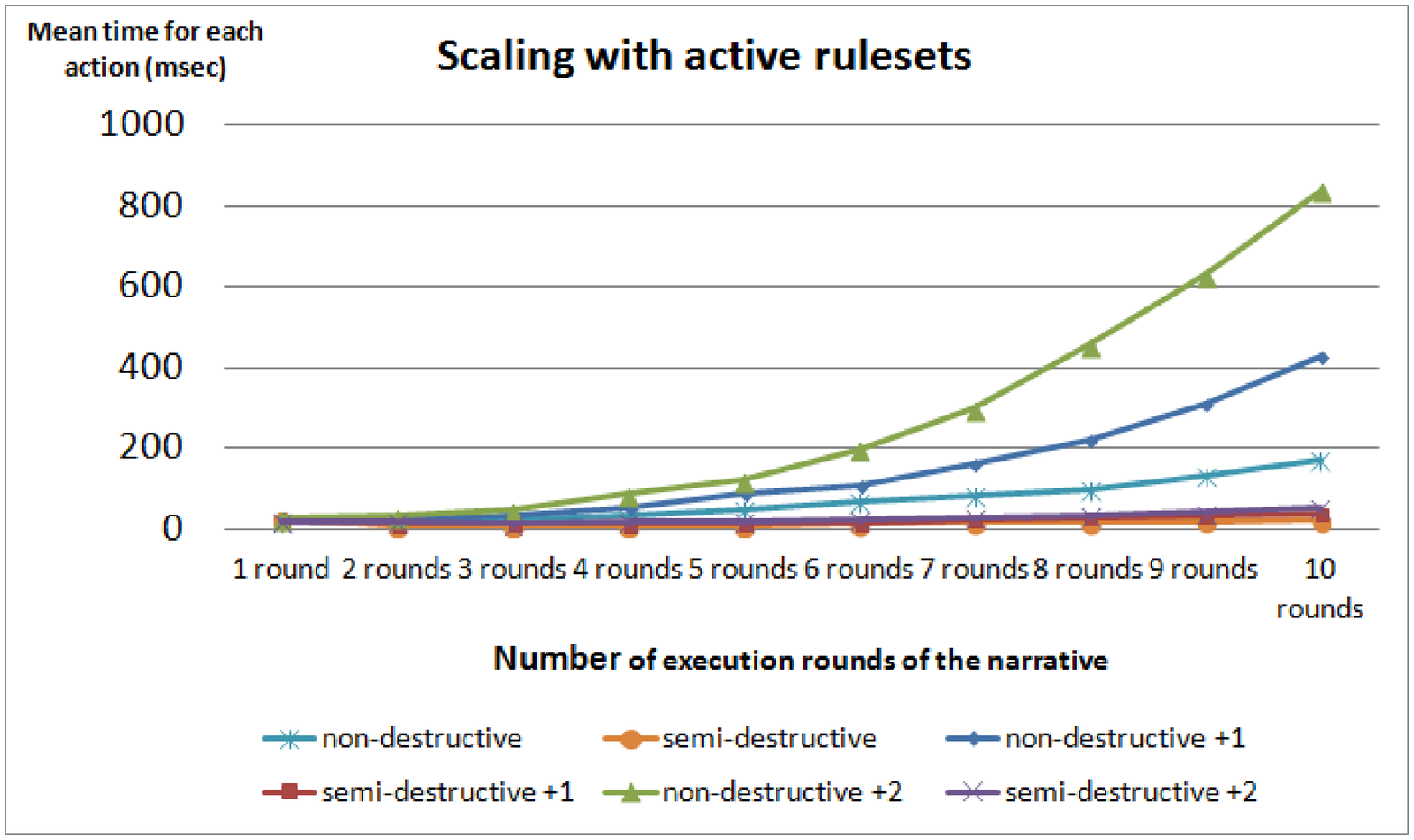}
\label{graph1b} }
\caption{Evaluation results.}
\end{figure}

\noindent \textbf{Scenario 1:}  Fig. \ref{graph1} displays the average reasoning time after each action within a given round, for successive executions of the same narrative. Specifically, at the beginning of each round two activities are considered as possible (two rulesets are activated), namely ``take shower'' and ``browse teeth'', but at the end only one is recognized with confidence. These activities take place in the bathroom and the only sensory information available to recognize them is the opening and closing of the hall-bathroom door - sometimes not even that, if for instance the door remained open from a previous interaction. For this reason, the rules that have been modeled to interpret the readings need to capture highly-expressive features, such as temporal relations, past activities, user habits etc..

The same narrative of actions is repeated in each round, in order for the results to be comparable. Moreover, we recorded both the non-destructive update of the KB, which maintains the value of all fluent, as well as the semi-destructive update. In the former case, the diagram displays also the number of facts stored in the KB. For example, after executing the same narrative 15 times, the KB contains 11885 facts in the non-destructive case and each action requires 567 msec on average to be processed. This means that the reasoner considers the full history of fluents to make a conclusion. In the semi-distructive case, on the other hand, only the current state of fluents is considered, leading to much better times ($<200$msec) (the history of events that have happened is kept in both cases, only the validity intervals of fluent is eliminated in the semi-destructive case).

\noindent \textbf{Conclusions:} What Fig. \ref{graph1} shows is that even after executing 20 rounds of typical user actions, keeping not only the recognized activities, but the validity interval of all fluents from the initial timepoint, response times are acceptable for the requirements of a practical system. Note that typically user activities are interdependent only in terms of sequence of executions and not with respect to the specific actions involved: it suffices to know during reasoning that the user has taken breakfast already, in order to deliberate whether he is preparing to take shower now (as extracted from the van Kasteren dataset), but not for how long he interacted with the groceries cabinet or the fridge while having breakfast. The non-destructive 20-round measurement assumes that all detailed information of the previous 19 rounds is included for reasoning, which is probably too demanding for real-world smart spaces. Usually, an activity can be recognized with confidence after a few rounds and old fluent intervals can be removed from the KB, thus maintaining reasonable sizes.

\noindent \textbf{Scenario 2:} The previous measurements investigated the performance of the system in the progress of time, as the user performs activities sequentially. In Fig. \ref{graph1b}, we concentrate on how the concurrent monitoring of \emph{potential} activities (axiomatized as different rulesets) affects performance. Imagine the case of a user entering the kitchen: he may get involved in numerous activities, some of which are irrelevant (e.g., making breakfast in the evening), while others get quickly discarded (e.g., putting dishes in the washing machine, unless certain evidences hold). The rest are potential activities that the user may start performing, whose confidence values need to be updated given the action narrative.

\noindent \textbf{Conclusions:} The diagram in Fig. \ref{graph1b} shows the response times with the same set of activities as before, as well as when one more and two more potential activities are added. To render the results comparable, the extra activities have the same types of rules (i.e., they are copies of the initial activities), yet they initiate completely different fluents. The results, in line with our previous analysis, reveal that it is the number of facts in the KB and the triggered rules that affect reasoning times: after 10 rounds, there are 9973 facts in the KBs with one extra activity resulting in 428 msec, and 13317 facts in the case of two extra activities, requiring 839 msec of reasoning (again times are significantly lower in the semi-destructive case).

As before, we expect the number of rounds needed to recognize with confidence an activity to be small in a real-world system, and the number of potential activities that need to be concurrently monitored to be limited, as well. In fact, this is one of the main benefits that our hybrid approach offers, as logic-based reasoning can be used to quickly eliminate irrelevant contingencies and then the probabilistic component can work more intensively on determining confidence values with more accuracy. Yet, all our diagrams present extreme cases, as well. It is  worth noting that the existence of activities (rulesets) that are not regarded as potential given a specific narrative has minimum effect on the performance of reasoning, as already made evident in Tables \ref{tableOBJ} and \ref{tableFL}, as well. Even with dozens of different activities being axiomatized, only the ones that the reasoner assumes as potentially occurring at a particular timepoint will affect reasoning times.

\section{Related Work}
\label{discussion}

Cerbere is a system for performing online reasoning in dynamic domains taking into account causal, temporal and epistemic notions. This article aims to demonstrate its potential for practical domains by framing its applicability in the context of smart spaces, a field that materializes a long anticipated objective for AI, due to the diversity and complexity of the challenges introduced.

For many years, the most popular tool to solve general Event Calculus problems has been the Discrete Event Calculus Reasoner \citep{Mueller06,Mueller04JLC}, which converts axiomatizations into satisfiability problems. The program implements the $\mathcal{DEC}$ variant and can support a wide range of commonsense features for automated deduction, abduction, postdiction and model finding tasks. Recently, ASP-based implementations have been proposed for very expressive fragments of the formalism, such as the one already mentioned by \cite{MillerMP13} that gives a Kripke-like semantics, in order to infer knowledge about domain fluents. Similarly, other logic programming implementations have been proposed, focusing on specialized problems. For instance, the Prolog-based implementation by \cite{ArtikisSP10} relies on the LTAR-EC dialect for recognizing activities and transforms effect axioms into a form that can effectively exploit Prolog's build-in indexing.

While these systems balance between expressiveness and efficiency, they do not support reactive features and they are generally not optimized for run-time execution, where streams of events arrive on-the-fly. This requires special care, as argued in related studies \citep{AnicicFodor10,ChesaniMelloMontali10}. The situation has started to change recently, with reactive Event Calculus implementations being proposed. For instance, \citet{Artikis12} implement the $\mathcal{RTEC}$ dialect, which aims to compute efficiently maximal intervals of fluents. The system developed achieves high performance for run-time event recognition, relying on a sliding window of computation that is maintained in working memory. Similarly, \citet{Bragaglia12} implement effect axioms within the Drools rule-engine and show that its performance is comparative to the fastest Prolog interpreter. Both systems present very promising results that are in line with our objectives, but are restricted in fragments of the calculus that do not support certain complex features, such as unknown fluents, non-deterministic effects or complex ramifications.
 
The systems by \citet{ChesaniMelloMontali10} and by \citet{KowalskiSadri10,KowalskiSadri12}, sharing a
similar objective, focus on combining both forward- and
backward-chaining rules and formally prove the properties of the
operational semantics. The former implements the $\mathcal{REC}$ dialect on top of the SCIFF framework, in order to introduce reactive features, but does not treat partial observability of the world state or non-determinism. This line of research has recently led to the noteworthy reformulation of the LPS framework \citep{KowalskiSadri15}, which succeeds to combine the reactive
rules and destructive updates of production systems, active databases and BDI languages with the logical representations and semantics of logic
programming, deductive databases and action theories in AI. 

\section{Conclusions}

In this article, we presented Cerbere, a reasoner that translates Event Calculus axiomatizations into production rules, managing to support a wide range of features for commonsense reasoning. A description of both theoretical and technical aspects that characterize the reasoner has been given. Moreover, we showcased the integration or Cerbere with a probabilistic component for supporting inference tasks in smart spaces. Its performance evaluation considered a domain-independent setting, as well as configurations of a typical smart space, revealing the applicability of the system in various conditions.

The development of Cerbere in the future will follow the progress achieved in the underlying formalisms, in order to accommodate further features. From the technical standpoint, we plan to investigate parallelization solutions, in order to exploit the progress in multi-core environments, so that each Jess instance is executed in a dedicated core. Of equal importance will be to evaluate the ease of writing rules with Cerbere, in order to axiomatize complex domains, as this will determine the acceptance and sustainability of the approach in practice.\\

\textbf{Acknowledgments} We wish to thank the anonymous reviewers for their insightful comments, their criticism and suggestions.


\appendix
\section{Extracts from the Resources to Aid the Reader}

\subsection{Cerbere Reasoner}
Cerbere as a standalone Event Calculus reasoner can be downloaded from:\\
\url{http://www.csd.uoc.gr/~patkos/tplpAppendix/Cerbere.rar}

To run Cerbere, one needs to copy the jess.jar file in the lib folder of the reasoner. This file can be downloaded from the Jess home page. We recommend Jess version 7.1p2 and JDK 7.

Source code for examples mentioned in the paper (more are available in the $/EC$ files folder of the reasoner):

\noindent - Ex1. Shanahan's circuit: An example of epistemic reasoning
\begin{itemize}
	\item source:\\
	\url{http://www.csd.uoc.gr/~patkos/tplpAppendix/ShanahanCircuit.ec}\\
	sample output:\\
	\url{http://www.csd.uoc.gr/~patkos/tplpAppendix/ShanahanCircuitOutput.txt}
\end{itemize}  

\noindent - Ex2. Multiple model generation:
\begin{itemize}
	\item source1:\\ \url{http://www.csd.uoc.gr/~patkos/tplpAppendix/MultiModels1.ec}\\
	sample snapshot:\\ \url{http://www.csd.uoc.gr/~patkos/tplpAppendix/MultiModels1Snapshot.JPG}
	
	\item source2:\\ \url{http://www.csd.uoc.gr/~patkos/tplpAppendix/MultiModels2.ec}\\
	sample snapshot:\\ \url{http://www.csd.uoc.gr/~patkos/tplpAppendix/MultiModels2Snapshot.JPG}
\end{itemize}

\subsection{Probabilistic Component}

Sample XMLs implementing Recognition and Composition eBNs are given next:
\begin{itemize}
	\item Recognition eBN for the Prepare Breakfast activity (XML):\\
	\url{http://www.csd.uoc.gr/~patkos/tplpAppendix/PrepareBreakfastBNRecognition.xml}
	
	\item Recognition eBN for the Take Shower activity (XML):\\
	\url{http://www.csd.uoc.gr/~patkos/tplpAppendix/TakeShowerBNRecognition.xml}
	
	\item Composition eBN for the Prepare Breakfast activity (XML): \\
	\url{http://www.csd.uoc.gr/~patkos/tplpAppendix/PrepareBreakfastBNComposition.xml}
\end{itemize}

\subsection{Domain-Independent Evaluation}

Execution codes to reproduce the results of Tables 2 and 3 can be found here (add/remove axioms to generate all cases shown in the tables): 
\begin{itemize}
	\item Experiment 1 with 1000 objects.\\
	\url{http://www.csd.uoc.gr/~patkos/tplpAppendix/exp1_1k_objects.ec}
	
	\item Experiment 2 with 1000 fluents.\\
	\url{http://www.csd.uoc.gr/~patkos/tplpAppendix/exp1_1k_fluents.ec}
\end{itemize}
 
\subsection{The Hybrid Framework}

\subsubsection{ Execution Instructions}

The framework is designed to run over the infrastructure installed at the Lissi lab that connects the reasoner with the underlying components. Yet, one can bypass the initialization phase and run executions of the hybrid framework with simulated events manually (i.e., events entered by the user, rather than retrieved from the sensors).

One can download the version of Cerbere that automatically uploads the necessary domain axiomatizations and eBN files for activity recognition from here:\\
\url{http://www.csd.uoc.gr/~patkos/tplpAppendix/CerbereHybrid.rar}\\
Please run using Eclipse: import as a new project, press run and wait 5 seconds, while the program tries to connect with the different components (connection will be unsuccessful of course, as the server and the actual sensors will not be running).

As before, one needs to copy the jess.jar file in the lib folder of the reasoner, before running. This file can be downloaded from the Jess home page. We recommend Jess version 7.1p2:\\
\url{http://www.jessrules.com/jess/download.shtml}

To enter simulated events, please do not use the Add Event or Observation button; this will only send the events to the reasoner, but it will not trigger the 3-step reasoning cycle of the hybrid framework. Instead, click inside the Eclipse Console and press Enter; a pop up will appear where you can enter the event, as if obtained from some sensor.

A user's manual, with sample executions can be downloaded from the following location:\\
\url{http://www.csd.uoc.gr/~patkos/tplpAppendix/User's%20Manual.pdf}

\subsubsection{Source Codes}

Our activity recognition framework relies on a set of Event Calculus axiomatizations that implement the rational behind our inference process. Below are some of the files uploaded to the Cerbere reasoner to recognitze activities, as well as to instruct proper actions based on them. Some of these axiomatizations are domain dependent, others are domain independent.

\textbf{Domain Definition}: The Basic Domain Axiomatization.ec defines the basic concepts that constitute the domain.\\
\textit{Definition of the basic concepts}:\\
\url{http://www.csd.uoc.gr/~patkos/tplpAppendix/basic%20domain%20axiomatization.ec}

\textbf{Monitoring of ADLs}:  Different .ec files can be created to model how the reasoning system should recognize the different activities of daily living of an individual, e.g., take shower axiomatization.ec, brush teeth axiomatization.ec, etc. Additional axiomatization files can be added by developers to monitor other activities, along with the actions that should be executed by the system in order to assist people during these activities.\\
\textit{Axiomatization of the Take Shower activity}:\\
\url{http://www.csd.uoc.gr/~patkos/tplpAppendix/take%20shower%20axiomatization.ec}\\
\textit{Axiomatization of the Brushing Teeth activity}:\\
\url{http://www.csd.uoc.gr/~patkos/tplpAppendix/brush%20teeth%20axiomatization.ec}

\textbf{Modeling of the reasoning system's behaviour}: The system.ec file implement the reasoner’s functionality. This is an important domain-independent file that models the behavior for Possible and Recognized Activities, and also implements the main reasoning steps. \\
\textit{Definition of the reasoner's functionality}:\\
\url{http://www.csd.uoc.gr/~patkos/tplpAppendix/system.ec}

\textbf{Spatial entities and their relations}: The spatial reasoning.ec axiomatizes spatial relations for the different entities of the domain. The spatial relations concerns the events and the space region in which they may occur.\\
\textit{ Definition of parameters for spatial reasoning}:\\
\url{http://www.csd.uoc.gr/~patkos/tplpAppendix/spatial%20reasoning.ec}

\subsubsection{How to reproduce the measurements of the experimental evaluation}

The version of Cerbere in the following link has hard-coded the narrative of actions that were used to run our experimental evaulation (please, follow the instructions in A.4.1):\\
\url{http://www.csd.uoc.gr/~patkos/tplpAppendix/Cerbere%20for%20experimental%20evaluation.rar}

To see the results after running the reasoner, open the "Output" tab and click on the root of the tree of models in the left panel, titled "Models produced". Statistical data for each execution step are presented, as shown in the snapshot here:\\
\url{http://www.csd.uoc.gr/~patkos/tplpAppendix/statistics.JPG}\\ Recall that we aggregate the reasoning times for each timepoint within every round, in order to calculate the mean time of each action (each of the 6 actions causes the reasoner to progress 3 timepoints, due to the 3-step cycle; thus, the first round lasts from timepoint 1 to 18, the second from timepoint 19 to 36 etc.).

Alternatively, one can manually add the actions that constitute one round. These are:
\begin{verbatim}
Happens(DoorOpens(Ned,HallBedroom,0), -1)
Happens(DoorOpens(Ned,HallBathroom,100), -1)
Happens(TriggerAlert(NoActivity,340), -1)
Happens(TriggerAlert(NoActivity,580), -1)
Happens(DoorOpens(Ned,HallToilet,636), -1)
Happens(DoorOpens(Ned,HallBedroom,650), -1)
\end{verbatim}
adding 15000 to the absolute time of each action after every round. 

\subsection{Event Calculus to Jess Parsing Methodology}

For each object, fluent or event type, such as
\begin{verbatim}
sort: object(O1,O2).
fluent: F1(object,object).
event: E1(object).
\end{verbatim}
one 'deffacts' Jess rule is created, with one tuple for each instance. For example:
\begin{verbatim}
(deffacts objects
(sort (name object)	(instance O1))
(sort (name object)	(instance O2)))

(deffacts fluentDEF_F1
(fluentDEF (name F1)	(argSort object object)))

(deffacts eventDEF_E1 
(eventDEF (name E1)	(argSort object)))
\end{verbatim}

These rules fire only once, during the initialization of the reasoner, producing all appropriate facts.

For each Event Calculus axiom, such as
\begin{verbatim}
Initiates(E1(?o2), F1(?o2, ?o1), ?t).
\end{verbatim}
two 'defrule' Jess rules are created. The idea is not to instantiate all fluents or events, until they are needed. Recall that Cerbere assumes that all fluents that do not exist in the KB are false. As such, we can manage the size of the KB in favor of efficiency. For the above axiom, we create:
\begin{verbatim}
(defrule EFFECTAXIOMS::0_AssertEffect_Initiates_E1_F1_?o2_?o1
(Time (timepoint ?t))
?event <- (event (name E1) (arg ?o2))
(EC (predicate Happens) (event ?event) (time ?t))
(fluentDEF (name F1) (argSort ?sort0 ?sort1))
(sort (name ?sort0) (instance ?o2))
(sort (name ?sort1) (instance ?o1))
(not (fluent (name F1) (arg ?o2 ?o1)))
=> 
(assert (fluent (name F1) (arg ?o2 ?o1) )))

(defrule EFFECTAXIOMS::1_Initiates_E1_F1
(Time (timepoint ?t))
?event <- (event (name E1) (arg ?o2))
(EC (predicate Happens) (event ?event) (time ?t))
?effect <- (fluent (name F1) (arg ?o2 ?o1))
=> (assert (EC (predicate Initiates)
(epistemic no)
(event ?event)
(posLtrs ?effect)
(time ?t)))) 
\end{verbatim}
The first rule instantiates the fluent and fires only once. The second fires every time the preconditions of the axioms are true.

This pattern is followed for any type of axioms, such as event occurrence axioms, effect axioms, observations, trigger axioms or other.
These Jess rules can become rather complicated, based on the preconditions of the axioms, such as whether negation is used, if value comparisons exist or if there are nested preconditions combining the above.
For instance, the axiom
\begin{verbatim}
HoldsAt(F1(?o1, ?o2),?t) ^
{?o1 <> ?o2} ^
~Happens(E2(O1),?t)=>
Initiates(E1(?o2), F1(?o2, ?o1), ?t).
\end{verbatim}
is parsed as the following Jess rule (only the one is shown here, the other is similar):
\begin{verbatim}
(defrule EFFECTAXIOMS::1_Initiates_posF1_negE2_E1_F1
(Time (timepoint ?t))
?event <- (event (name E1) (arg ?o2))
(EC (predicate Happens) (event ?event) (time ?t))
?effect <- (fluent (name F1) (arg ?o2 ?o1))
?f0 <- (fluent (name F1) (arg ?o1 ?o2))
(EC (predicate HoldsAt) (epistemic no) (posLtrs ?f0 ) 
    (time ?tf0&:(or (eq ?tf0 ?t) (eq ?tf0 -1)) ))
(test (neq ?o1 ?o2))
(not (and	?e2 <- (event (name E2) (arg O1))
(EC (predicate Happens) (event ?e2rxl ) (time ?t))
(test (eq ?e2 ?e2rxl))
))
=> (assert (EC (predicate Initiates)
(epistemic no)
(event ?event)
(posLtrs ?effect)
(time ?t))))
\end{verbatim}
Notice how all variables are instantiated in the body of the Jess, how the condition between variables is translated into a 'test' Jess command and how the event occurrence precondition exists inside a 'not' statement.




\newpage


\bibliographystyle{acmtrans}

\bibliography{TPLP14Cerbere}

\end{document}